\documentclass{article} 
\usepackage{natbib}
\usepackage{iclr2024_preprint,times}
\usepackage{colortbl}  

\usepackage[pagebackref,breaklinks,colorlinks=true,citecolor=blue]{hyperref}
\usepackage[addedmarkup=uwave,defaultcolor=red]{changes}

\usepackage{url}
\usepackage{tcolorbox}
\usepackage{graphicx}
\usepackage{makecell}
\usepackage{framed}
\usepackage{amsmath}
\usepackage{enumitem}
\usepackage{pifont}
\definecolor{shadecolor}{rgb}{0.92,0.92,0.92}
\usepackage{wrapfig}
\usepackage{booktabs}
\usepackage{amsfonts}       
\usepackage{nicefrac}       
\usepackage{microtype}      
\usepackage{xspace}
\usepackage{makecell}
\usepackage{multirow}
\usepackage[capitalize]{cleveref}
\usepackage{scalerel}
\usepackage[misc]{ifsym}

\makeatletter
\DeclareRobustCommand\onedot{\futurelet\@let@token\@onedot}
\def\@onedot{\ifx\@let@token.\else.\null\fi\xspace}

\def\ie{\emph{i.e}\onedot}

\def\etc{\emph{etc}\onedot}

\makeatother

\makeatletter
\renewcommand{\paragraph}{%
  \@startsection{paragraph}{4}%
  {\z@}{0ex \@plus 0ex \@minus 0ex}{-1em}%
  {\hskip\parindent\normalfont\normalsize\bfseries}%
}
\makeatother

\def\emoji{\scalerel*{\includegraphics[height=0.09in]{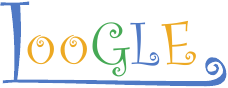}}{\textrm{\textbigcircle}}}
\newcommand{\logo}{\emoji{}\xspace}

\newcommand{\cmark}{\ding{51}}%
\newcommand{\xmark}{\ding{55}}%

\usepackage{amsmath,amsfonts,bm}

\def\eqref#1{equation~\ref{#1}}

\def\1{\bm{1}}

\DeclareMathAlphabet{\mathsfit}{\encodingdefault}{\sfdefault}{m}{sl}
\SetMathAlphabet{\mathsfit}{bold}{\encodingdefault}{\sfdefault}{bx}{n}

\crefname{table}{Tab.}{Tabs.}

\title{\logo: Can Long-Context Language Models Understand Long Contexts?}

\author{
Jiaqi Li$^{1,}$\thanks{Equal contributions.} , Mengmeng Wang$^{1,*}$, Zilong Zheng$^{1,}$\thanks{Correspondence to Zilong Zheng \texttt{<zlzheng@bigai.ai>} and Muhan Zhang \texttt{<muhan@pku.edu.cn>}.} , Muhan Zhang$^{1,2,\dagger}$ \\
$^1$ National Key Laboratory of General Artificial Intelligence, BIGAI \\
$^2$ Institute for Artificial Intelligence, Peking University
}

\newcommand{\dataset}{\textbf{LooGLE
}\xspace}

\iclrfinalcopy 
\begin{document}

\maketitle

\begin{abstract}

Large language models (LLMs), despite their impressive performance in various language tasks, are typically limited to processing texts within context-window size. This limitation has spurred significant research efforts to enhance LLMs' long-context understanding with high-quality long-sequence benchmarks. However, prior datasets in this regard suffer from shortcomings, such as short context length compared to the context window of modern LLMs; outdated documents that have data leakage problems; and an emphasis on short dependency tasks rather than long dependency tasks.
In this paper, we present \logo, a \textbf{Lo}ng C\textbf{o}ntext \textbf{G}eneric \textbf{L}anguage \textbf{E}valuation benchmark for LLMs' long context understanding. 
\logo features relatively new documents post-2022, with over 24,000 tokens per document and 6,000 newly generated questions spanning diverse domains. Human annotators meticulously crafted more than 1,100 high-quality question-answer pairs to meet the long dependency requirements. These pairs underwent thorough cross-validation, yielding the most precise assessment of LLMs' long dependency capabilities. The evaluation of eight state-of-the-art LLMs on \logo revealed key findings: (i) commercial models outperformed open-sourced models; (ii) LLMs excelled in short dependency tasks like short question-answering and cloze tasks but struggled with more intricate long dependency tasks; (iii) in-context learning and chaining thoughts offered only marginal improvements in long context comprehension; (iv) retrieval-based techniques demonstrated substantial benefits for short question-answering, while strategies for extending context window length through optimized transformer architectures or positional encoding had limited impact on long context understanding. As such, \logo not only provides a systematic and comprehensive evaluation schema on long-context LLMs, but also sheds light on future development of enhanced models towards ``true long-context understanding''.
All evaluation codes are released at: \url{https://github.com/bigai-nlco/LooGLE}.
\end{abstract}

\begin{figure}[h!]
    \vspace{-0.3cm}
    \centering
    \includegraphics[width=\linewidth]{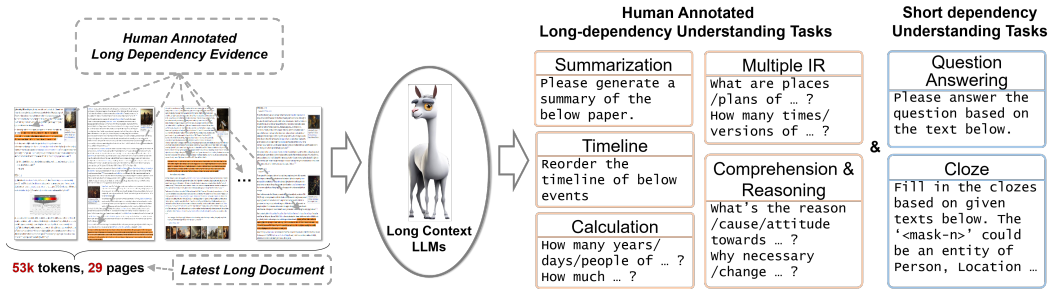}
    \caption{The LooGLE benchmark for long context understanding.}
    \label{fig:overview}
\end{figure}


\section{Introduction} 
\vspace{-0.9em}
The pursuit of enabling large language models (LLMs), such as ChatGPT~\citep{brown2020language, openai2023gpt4, zeng2023matters}, to go beyond their limited context window size so as to process, comprehend, or even learn from long-context textual information~\citep{ding2023longnet, dao2022flashattention, chi2023dissecting, bulatov2023scaling} is inevitable for next-generation of language intelligence attributed to its wide applications on real-world scenarios, such as domain-specific knowledge understanding, long-context conversational generation, long story or code generation, \etc.

Meanwhile, there is an increasing need for high-quality benchmarks with much longer text lengths and more challenging tasks to provide comprehensive evaluations. However, traditional benchmarks~\citep{cohan2018discourseaware, sharma2019bigpatent, huang2021efficient} often fall short in text length with an average number of thousand words~\citep{narrativeqa, yang2018hotpotqa}. Besides, existing benchmarks automatically collect possibly outdated documents from existing datasets published a few years ago~\citep{shaham2022scrolls, trivedi2022musique, wang2022squality, angelidis2020extractive}, which might lead to data leakage in pre-trained LLMs and make the evaluation inaccurate. Further, the long texts are often restricted to domain-specific articles, making it hard to evaluate LLMs' ability on generic tasks and domains. Finally, it is important to note that tasks in existing benchmarks are primarily \textbf{short dependency} tasks, which only require LLMs to retrieve answers from one specific sentence or paragraph, without really testing LLMs' ability to collect pieces of information from paragraphs across the whole document and summarize them into an answer, which we call \textbf{long dependency} tasks.

\begin{table}[t!]
    \vspace{-10pt}
    \centering
    \caption{Comparison with other long-context benchmarks.}
    \label{tab:benchmark_comparison}
    \resizebox{\linewidth}{!}{%
    \begin{tabular}{lccccccccc}
    \toprule
        \multirow{2}{*}{\textbf{Dataset}} & \multirow{2}{*}{\textbf{Avg. Words}} & \multirow{2}{*}{\textbf{\# of Docs.}} & \multirow{2}{*}{\textbf{\# of Ques.}} & \multirow{2}{*}{\textbf{Manually Label}} & \multicolumn{5}{c}{\textbf{Long Dependency Tasks}}\\ \cmidrule{6-10}
        &  & & & & Summarization & Info. Retrieval & Timeline Reorder & Computation & Doc QA \\  
        \midrule
        Zero Scrolls~\citep{shaham2023zeroscrolls} & 10,392 & - & 4,378 & \xmark & \cmark & \xmark & \xmark & \xmark & \xmark\\
        Long Bench~\citep{bai2023longbench} &8,120 & - & 4,750 & 350 & \cmark & \cmark$^*$ & \xmark & \cmark$^*$ & \cmark\\
        L-Eval~\citep{an2023leval} & 8,008 & 411 & 2,043 & 2,043$^\dagger$ & \cmark & \cmark & \xmark & \xmark & \cmark\\ \midrule
        \dataset~(Ours) & 19,367 & 776 & 6,448 & 1,101 & \cmark & \cmark &\cmark &\cmark &\cmark \\
    \bottomrule
    \multicolumn{10}{l}{\footnotesize $^*$ The task is created in a synthetic manner.} \\
    \multicolumn{10}{l}{\footnotesize $^\dagger$ The questions are re-labelled from original data.} \\
    \end{tabular}
}
\end{table}

To mitigate the shortcomings of existing datasets, in this paper, we introduce a novel benchmark \dataset, short for \textbf{Lo}ng C\textbf{o}ntext \textbf{G}eneric \textbf{L}anguage \textbf{E}valuation, to evaluate the long context understanding abilities of LLMs illustrated in \cref{fig:overview}. Our benchmark has the following advantages:
\begin{itemize}[leftmargin=*,topsep=0pt,noitemsep]
    \item \textbf{Extra-long realistic documents}. It contains 776 latest gathered and extremely long documents with an average of 19.3k words. There are over 6,448 test instances without distribution bias for a more generalized assessment, many of which exceed 100k words. On one hand, they can better evaluate LLMs' capability on memorizing and understanding longer text that is far beyond their context window size. On the other hand, the excessive length is well suited to the common usage of long text scenarios.
    \item \textbf{Manually designed both short and long dependency tasks.} It is composed of 7 major tasks to evaluate LLMs' ability to understand both short and long dependency content. We refer ``long dependency" tasks as those that require the understanding of the inter-dependency across multiple evidence widely spanning over the entire long text. We delicately design 5 types of long dependency tasks and recruited a group of human annotators to manually create 1101 long dependency Question-Answer (QA) instances, despite the high costs and huge effort involved in this process.
    \item \textbf{Relatively new documents.} Our benchmark comprises texts all published after 2022 which ensures that most modern LLMs (at the date of submission) have not been pre-trained on these documents, \textbf{forcing them to rely on their in-context learning ability} rather than memorization. In contrast, existing benchmarks are usually a combination of content from traditional NLP dataset, whose world knowledge may have already been learned by LLMs and thus are less convincing for assessment. Furthermore, our data collection process is fully open-sourced, making it easy for the community to reconstruct/update the benchmark with newer documents, possibly on a yearly basis.
    \item \textbf{Cross-domain generic data.} Our benchmark is derived from popular open-source documents, including arXiv papers, Wikipedia articles, and movie and TV scripts, spanning diverse domains and multiple categories such as academia, history, sports, politics, arts, events, and entertainment. 
\end{itemize}

We conduct a comprehensive evaluation of 8 representative LLMs on \dataset. We specifically select LLMs which have made great effort in addressing the challenge of understanding long contexts as the baselines. The results indicate that better base models with a larger context window size generally achieve better performance. However, all models experience a significant performance decline in long dependency tasks, indicating there is a desperate need to improve the true long dependency understanding capabilities of LLMs. Our dataset serves as an up-to-date benchmark for cutting-edge assessment and research on the long context understanding and modeling of LLMs.


\section{Related Work}
\paragraph{Existing models for long context understanding.} There are increasing research interests in developing methods to extend LLMs' context window size, such as utilizing
recurrent memory, sparse attention~\citep{meister2021sparse}, external memory and etc.\citep{chen2023longlora, xiong2023effective,li2023functional}. The most popular way is to develop improved transformer architectures~\cite {dong2023survey}. Efficient transformers~\citep{tay2020long, tay2022efficient} are proposed to decrease the memory and time complexity to efficiently model longer texts. Unlike efficient transformers that simplify the attention structure, recurrent transformer~\citep{bulatov2022recurrent,bessonov2023recurrent} keeps the full self-attention mechanism. History information of previous segments is cached and will be leveraged when the subsequent segment is fed into the model without a context fragmentation problem. Fine-tuned models on long documents~\cite {wu2021recursively} are also explored, but they are often effort-costing and face difficulties in collecting ground truth fine-tuning data for long text tasks. Apart from approaches which are developed from modeling and parameter updating aspects, there are also works incorporating external memory structures and compression techniques for LLMs or using task-oriented process optimization strategies~\citep{gidiotis2020divideandconquer, zhou2022finegrained, ram2023incontext, izacard2022atlas}.

\paragraph{Existing datasets for long context understanding.}
There are a growing number of benchmarks proposed to test LLMs' long context understanding ability~\citep{shaham2023zeroscrolls, li2023unlocking}. ZeroSCROLLS, L-Eval and LongBench are the three most recent ones. ZeroSCROLLS~\citep{shaham2023zeroscrolls} automatically processes datasets from different sources into a unified input format with an average of 10k words. However, it mainly focuses on collecting documents and tasks from existing datasets and relies on automatic metrics for limited model comparisons~\citep{shaham2022scrolls}. L-Eval~\citep{an2023leval} differs in re-annotating the data and instructions from similar public datasets with smaller sizes to ensure the quality. Besides, it optimizes the evaluation procedures and baselines to get more accurate conclusions. LongBench~\citep{bai2023longbench} provides a bilingual and multi-task dataset featuring diverse sequences of varying lengths, distributions, patterns, languages and domains for a comprehensive evaluation of long context understanding. Nonetheless, it encompasses texts of only thousands of words and tasks mostly restricted to short-term information extraction. Moreover, there are few types of ``long dependency" tasks in previous datasets, except for summarization (which LLMs are validated to perform well on) and synthesized tasks like data aggregation and retrieving. To finish those tasks, LLMs solely need to locate pieces of information from the lengthy source input and aggregate them together. In contrast, we propose \dataset which contains long dependency tasks that are much more challenging, such as event timeline reordering, comprehension/reasoning, and computation. These tasks require not only information retrieval but also understanding/reasoning over the entire text. We include a detailed comparison with concurrent works in \cref{tab:benchmark_comparison}.

\section{The \dataset Benchmark}
There are three categories of data sources as mentioned in \cref{tab:benchmark}. Based on that, we generate two main types of tasks: short dependency and long dependency tasks in \dataset. For short dependency tasks, we generate short QA from Wikipedia articles and cloze from scripts. For the long dependency tasks, we include summarization for arXiv papers and manually designed QA tasks for long document understanding. There are four major subtasks for QA: Multiple information retrieval, Timeline reorder, Computation, Comprehension and reasoning. We delicately generate tasks/questions to customize the intrinsic features of each data source for better long-context understanding assessments.

\label{headings}

\subsection{Dataset selection and construction}
Our \dataset benchmark consists of 3 sources: scientific papers, Wikipedia articles, movie and TV scripts, all covering various topics and categories. These documents are commonly used as corpora in NLP tasks. By replicating the methodology proposed in this paper, they can be collected easily and periodically. All the documents in our \dataset benchmark are after 2022 and filtered by a length of over 10k words. We have also considered books, but found that most books meeting our principles are not license-free, therefore giving them up. Statistics of the three sources can be found in \cref{tab:benchmark}. Details of the dataset are introduced in the following sections.

\textbf{arXiv papers}
\quad 
We pulled data from a massive pool of 10,000 entries on the arXiv website (https://arxiv.org/) using a random selection method. These entries ranged from January 2022 to April 2023. In the next step, we extracted their abstracts, making them our main source for the summarization task. We were pretty rigorous about maintaining data quality. That meant ditching the reference sections, cleaning up any garbled characters from math equations, and leaving out any documents under 10,000 words. After all that thorough check, we ended up with a solid collection of 516 research papers.

\begin{table*}[t!]
\centering
\resizebox{\linewidth}{!}{%
\renewcommand{\arraystretch}{0.9}

\begin{tabular}{lccccccc}
\toprule
{\textbf{Dataset}} & 
{\textbf{No. Docs}} &
{\textbf{Avg. Words}} &
{\textbf{Max. Words}} &
{\textbf{Min. Words}} &
{\textbf{Avg. Tokens}} &
\textbf{Task} &  
{\textbf{\# Questions}}  \\ 
    \midrule

arXiv   & 516                  & 16,988     & 197,977   & 10,204               & 20,887                           & Summarization     & 516  \\ \midrule
\multirow{2}{*}{Wikipedia}     & \multirow{2}{*}{105} & \multirow{2}{*}{17,604} & \multirow{2}{*}{46,250} & \multirow{2}{*}{11,285} & \multirow{2}{*}{21,017} & Short dependency QA                         & 1,951 \\
    &  & &  &  & & Long dependency QA  & 459  \\ \midrule

\multirow{2}{*}{\makecell[l]{Movie \& \\ TV scripts}} & \multirow{2}{*}{155} & \multirow{2}{*}{28,483}  & \multirow{2}{*}{62,752}  & \multirow{2}{*}{11,089} & \multirow{2}{*}{36,412} & Cloze & 2,880  \\
    &  & &  &  & & Long dependency QA  &  642 \\ 
  \bottomrule
\end{tabular}
}
\caption{\textbf{Statistics of LooGLE}}\label{tab:benchmark}
\end{table*}

\textbf{Wikipedia articles}
\quad Wikipedia is a free and popular online encyclopedia that provides information and reference on a wide range of topics. Articles are created and edited collaboratively by volunteers from all around the world, making it a dynamic and constantly evolving resource. These Wikipedia articles are perfect for evaluating the long text reading, comprehension, summarization, and information retrieval abilities of LLMs. We first downloaded and parsed the most recent page articles present in .bz file format from the official website (https://dumps.wikimedia.org/). Then we kept the articles after 2022 with over 10k words utilizing a subset of the open-source Wikipedia dataset (202203.en) from Hugging Face (https://huggingface.co/datasets/wikipedia). Since some pages in the dump file probably no longer exist and are redirected to a relevant page, we only retain pages (exempt summary, citations and references) after redirection.

\textbf{Movie and TV scripts}
\quad A movie or TV script typically contains essential information such as scene descriptions, action descriptions, and dialogues between characters. Scripts inherently encapsulate numerous events and facts in dialogue format, necessitating models to deeply comprehend contextual nuances. To comprehend the events unfolding within a dialogue, there is a high demand on reasoning ability, along with the ability to navigate shifts in perspective and grasp the viewpoints of the characters involved. Additionally, scripts are typically lengthy and challenging for LLMs with fixed context window sizes. All scripts are sourced from three websites (https://www.scriptslug.com, https://thescriptlab.com/, https://8flix.com), consisting of movies and TV shows released after 2022.

\subsection{Long dependency tasks}

\subsubsection{Tasks definition}

\paragraph{Summarization}
We directly use the abstract of each paper as the reference for generating summaries. The abstracts effectively capture the main content and key information of each paper.

\paragraph{Long dependency QA}
One highlight of our dataset is that we dedicated significant effort to manually compile about 1.1k true long dependency QA pairs. The construction process is detailed in the next section. We manually designed 4 long dependency tasks:  Multiple information retrieval, Timeline reorder, Computation, Comprehension and reasoning. As we will show in the experiments, these tasks are pretty challenging, requiring more advanced capabilities for long context understanding. They are valuable for understanding the limitations of LLMs. Examples of the 4 types of long dependency QAs are shown in \cref{fig:longQA}.

\begin{itemize}[leftmargin=*,topsep=0pt,noitemsep]
\item  \textbf{Multiple information retrieval:} Quite different from traditional short-term retrieval tasks, there are usually multiple and diverse pieces of evidence throughout the entire text for one specific answer. 
The task requires extensive information extraction from widely distributed segments within the lengthy text, followed by the aggregation of the evidence to derive the ultimate answer. The evidence is distinctly presented and can be directly located within the original sentences or sections of the text.

\item  \textbf{Computation:} Similar to the previous task, it firstly needs multiple information retrieval from a wide range of texts. 
A majority of the evidence within the text takes the form of numerical data, often in question formats such as inquiries about quantities, frequencies, durations, specific numbers, and so on. To arrive at an accurate response, a profound comprehension of the question and its correlation with the provided numerical data is essential. This process relies heavily on the capacity to grasp extensive contextual information and also involves a degree of mathematical reasoning ability.

\item  \textbf{Timeline reorder:} 
This task follows a more conventional format, involving the instruction, ``Please reorder the timeline of the following events,'' along with a set of events presented in a permuted order. The objective is to arrange these events in accordance with their chronological sequence as dispersed throughout the extensive text. The events are derived directly from the source text, either as extracted segments or summarized factual information. Successful completion of this task necessitates either the memorization or comprehensive understanding of the central storyline of the document and assesses the model's proficiency in temporal awareness.

\begin{figure}[t!]
    \centering
    \includegraphics[width=0.85\textwidth]{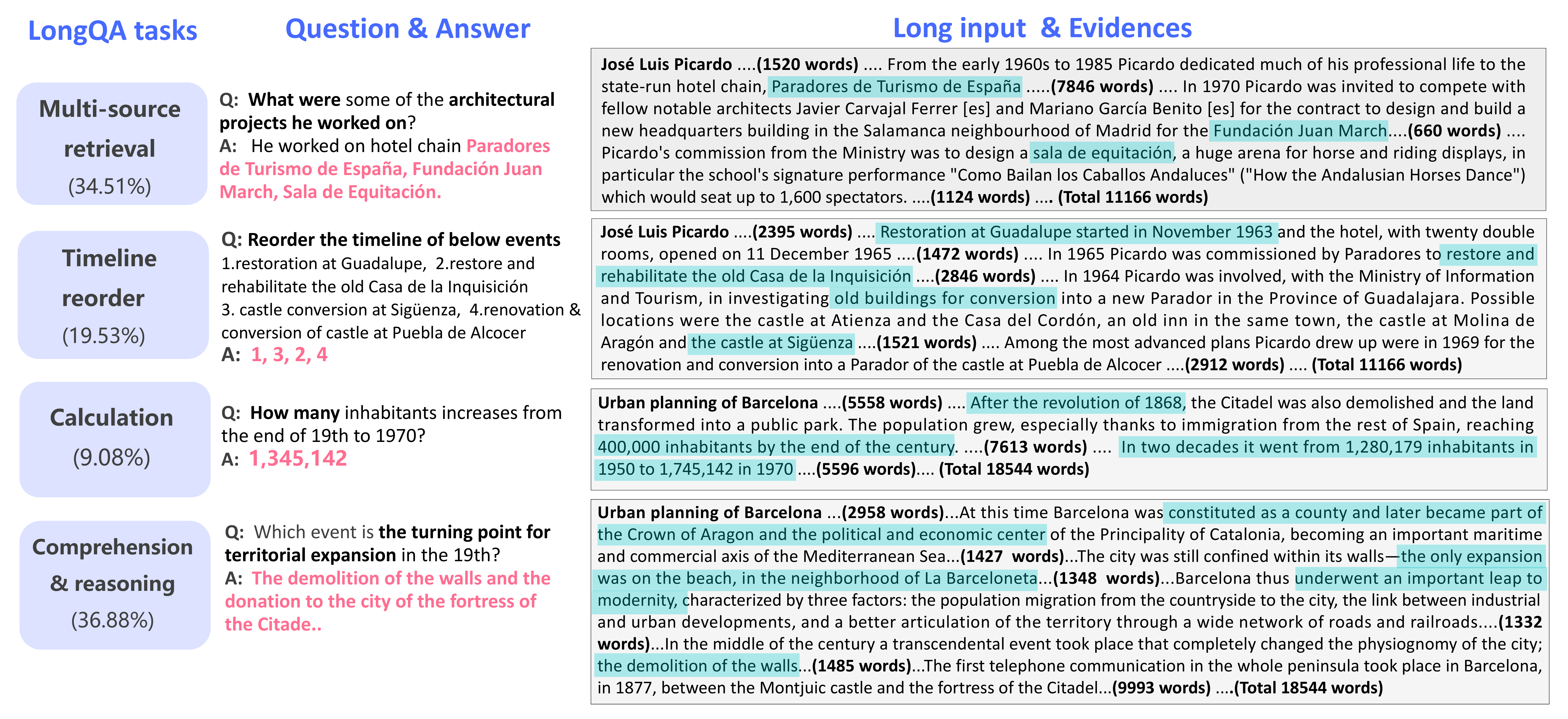}
    \vspace{-10pt}
    \caption{Long dependency QA tasks}
    \label{fig:longQA}
\end{figure}

\item \textbf{Comprehension and reasoning:} 
This task demands not only a profound comprehension of the question but also intricate reasoning to discern the underlying implications for searching for the appropriate evidence. The most prevalent question patterns involve inquiries about causality, impact, contributions, attitudes, and the essential attributes related to various events. Additionally, more extensive comparisons and evaluations are essential when the questions revolve around the primary, predominant, highest, or most critical aspects of the evidence. Furthermore, the answers to this task are not explicitly evident within the source text. They often require multi-step reasoning to model the inherent connections and dependencies, facilitating the acquisition of the answer through a complex analytical process.
\end{itemize}

\subsubsection{Construction process of long dependency QAs}
We detail the construction process as follows. We first randomly sampled a total of 140 long documents from Wikipedia and the scripts dataset. We recruited students from top universities across the nation and organized a manual annotation process to generate long dependency QAs. We categorize long dependency tasks into Multiple information retrieval, Comprehension and reasoning, Calculation, and Timeline reorder (illustrated in \cref{fig:longQA}). Each document spans from 10,000 to 20,000 words in average and requires a generation of 5 to 10 questions. Additionally, participants were prohibited from employing large language models and tools like ChatGPT for article reading, data generation, and annotation.

In the generation of questions, each document underwent a meticulous three-step process that involved the assignment of two distinct annotators --- one serving as the questioner and the other as the answerer. Importantly, these annotators were kept unaware of each other's identities, ensuring a rigorous cross-validation process to maintain the quality of the questions, answers, and supporting evidence. This approach aimed to achieve questions with a high degree of accuracy, precision, and relevance to the document's content.

\textbf{Step 1: Question and answer.} 
The questioner's role encompassed a comprehensive set of responsibilities, including reading the document, crafting relevant questions, offering their own answers to those questions, and pinpointing the specific evidentiary passages within the document that substantiated their answers.
The annotation adhered to stringent standards, encompassing the following key principles:
\begin{itemize}[leftmargin=*,topsep=0pt,noitemsep]
\item Long dependency: Each question was required to exhibit a long dependency, \ie, the evidence supporting its answer should have a wide span across the document. The recommended dependency length (the distance between the earliest and latest evidence) is a minimum of 5,000 words.
\item Diverse problem types: The questioner was required to generate a set of 5 to 10 question-answer pairs for each document, which should not contain more than 4 questions of the same type to avoid imbalanced question distribution and prevent annotators from generating overly simple questions.
\item Clear and precise questions: The formulation of each question was asked to adhere to clarity, conciseness, and no ambiguity, with examples provided.
\item Deterministic and objective answers: The answers to the proposed questions were rigorously checked to be deterministic and objective, precluding open-ended ones.
\end{itemize}
\textbf{Step 2: Answer and check.} The second step involves the answerers. Each answerer can only access the assigned article text and the posed questions from the questioner in the first step. The answerer was required to thoroughly read the entire document and provide answers to the questions accordingly. The standard for the answers is the same as the questioners. In addition to the aforementioned responsibilities, the answerer was also tasked with assessing the quality of the questions, which entails evaluating whether the questions adhere to the standard and whether they are answerable. In instances where a question cannot elicit a definite and unambiguous answer, it is deemed as unsatisfactory, and the answerer is asked to provide constructive feedback for improvement. 

\textbf{Step 3: Revise.} In the third step, the questioner for the document had access to the document, the questions, the two sets of answers from both the questioner and the answerer, as well as the feedback from the answerer. The questioner was asked to first revise the questions according to the feedback, and then unify their own answers with those from the answerers to derive the final answers.

In the first step, we acquired a total of 1,137 question-answer pairs. In the second step, 206 of these pairs were identified as non-compliant with the established criteria and were accompanied by suggestions for improvement. The inter-annotator agreement rate is 81.88\%~\citep{kim2023interannotator}. Following the revisions conducted in the third step, we ultimately obtained a total of 1101 high-quality long dependency question-answer pairs which require strong long context understanding ability.

\subsection{Short dependency tasks} 

\paragraph{Question Answering (QA)}
To generate short dependency QA pairs, we harnessed the robust language processing and comprehension capabilities of GPT3.5-turbo-16k. These short dependency QA pairs typically do not require extensive evidence retrieval and can be extracted from localized segments. We divided each article into multiple segments and employed an iterative approach to prompt the Language Model (LLM) to generate QA pairs based on these segments, including their associated supporting evidence from the article. Details of the prompts are available in Appendix~\ref{appendix:prompts}. Subsequently, we conducted manual reviews of the QA pairs, making refinements to some of the answers by filtering out non-essential context and eliminating redundant descriptions. This rigorous curation process was undertaken to ensure the high quality and relevance of the resulting QA pairs.

\paragraph{Cloze}  Initially, each script is divided into segments of varying lengths. Then, we employ GPT3.5-turbo-16k to generate factual summaries aligning with the source segment along with some constraints included in prompts (see Appendix~\ref{appendix:prompts}). Later, we employ BERT-large ~\citep{devlin2019bert} for Named Entity Recognition (NER)~\citep{roy2021recent} from the generated summaries, limiting the types to person name, location, and organization. Finally, we randomly select a certain number (no more than 5) of entities from the summary and mask them as placeholders, denoted as ``\textless mask-n\textgreater". The goal is to predict the masked entities according to the long context. 

\section{Evaluation}

\subsection{Models Selected for Evaluation}

\paragraph{Commercial models}  GPT4-32k, GPT4-8k, GPT3.5-turbo-16k~\citep{chen2023robust,ye2023comprehensive} are all the models developed by OpenAI, as documented on their official platform  (https://platform.openai.com/docs/models). GPT4-32k can handle up to 32k tokens in the context input, and GPT4-8k and GPT3.5-turbo-16k can handle up to 8k and 16k context input, respectively. We use the models of version 0613 by default.

\paragraph{Open-source models} LLaMA2-7B-32K~\citep{touvron2023llama} is developed by Together (https://together.ai/) and fine-tuned from Meta's original Llama2-7B ~\citep{touvron2023llama}  model. It has been expanded to accommodate a context length of 32K using Position Interpolation~\citep{chen2023extending}. ChatGLM2-6B-32k~\citep{du2022glm}, is a product of THUMD and represents an enhancement of the ChatGLM2-6B model. It is notable for its integration of FlashAttention ~\citep{dao2022flashattention}, allowing it to train with an extended context length, increased from 2K to 32K. LongLLaMa-3B, derived from openllama, has been fine-tuned using Focused Transformer ~\citep{tworkowski2023focused} to extend its context to 256k. Lastly, RWKV-4-14B-pile ~\citep{peng2023rwkv} is a member of the RWKV model family, notable for its architectural fusion of both Recurrent Neural Networks (RNN) and Transformers. It has been fine-tuned to accommodate a context length of 8K.

\paragraph{Retrieval-based Method} Instead of extending the context window size, retrieval-based context compression technique~\citep{xu2023retrieval,askari2023retrieval} augments the LLM by incorporating external memory, allowing relevant information to be retrieved using a specific query. LlamaIndex (https://github.com/jerryjliu/llama\_index) is a data framework designed for LLMs. It fulfills a dual role by constructing indices for document segments and functioning as an intermediary connecting LLM with data sources, which enables LlamaIndex to retrieve relevant data segments before they are input into the LLM, thereby enhancing the LLM's capacity to effectively handle lengthy text. In our experiment, we employed the default configuration of the LlamaIndex, with embedding model text-embedding-ada-002 (https://openai.com/blog/new-and-improved-embedding-model) and language model text-davinci-003 
~\citep{ouyang2022training}.

It has been proved that, performance is often highest when relevant information occurs at the beginning or end of the input context, and significantly degrades when models must access relevant information in the middle of long contexts. \citep{liu2023lost}. Therefore, we artificially truncate the input document to certain sizes (all not larger than the context window size of above mentioned models) by concatenating the head and tail of the input. For example, when we want to truncate a long document to 16k, we concatenate its head 8k tokens and tail 8k tokens before feeding it to an LLM.

\begin{figure}[t!]
    \centering
    \includegraphics[width=1.01\textwidth,height=0.31\textwidth]{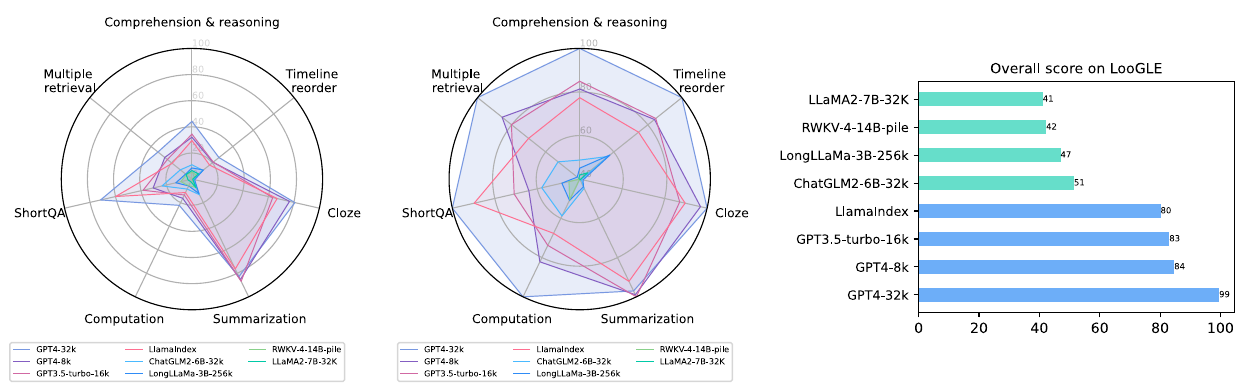}
    \caption{An overview performance of LLMs on \dataset for long context understanding}
    \label{fig:performance}
\end{figure}

\subsection{Evaluation methods and metrics} 

\paragraph{Automatic evaluation} 
We adopt several automatic evaluation metrics, which can be categorized into two types. Bleu, Rouge, Meteor Score and Bert Score~\citep{li2023large, mukherjee2023employing} are widely used for generative tasks such as summarization and QA. They evaluate the matching between groundtruth and LLM answers mainly based on n-gram matching and semantic similarity. For Cloze, Exact Match and Partial Match~\citep{sharma2023augmenting,engelbach2023finetuning} are employed in our evaluation. Exact Match entails the predicted entity and the groundtruth entity exactly match each other while Partial Match allows for fuzzy matching.

\paragraph{GPT4-as-judgment} 
Most automatic evaluation metrics are sensitive to semantic expression, output format, and length. Thus, these metrics alone might be insufficient for effectively comparing different models (some models might output answers in a style more similar to groundtruth). However, recent research has shown that the GPT4 evaluator exhibits high consistency with human evaluation and can serve as a reliable annotator to some extent~\citep{suri2023large,liu2023calibrating,zheng2023judging}. To provide a more comprehensive assessment of models, we utilize GPT4-8k as an LLM evaluator. For QA task, given one question and two answers provided by the groundtruth and the LLM's prediction, we ask GPT4-8k to judge whether the two answers are semantically the same or not. Then we calculate the accuracy that LLM answers match the groundtruth. For summarization task, given the predicted summary with the goundtruth, we ask LLM to give a score considering various factors for generation. The prompts implemented can be found in Appendix~\ref{appendix:prompts}.

\paragraph{Human evaluation} 
We also include human evaluation for reference, where we manually check whether LLM's prediction matches the groundtruth.

\begin{table}[t!]
\caption{Performance of the short dependency tasks}
\label{tab:short_term_qa_performence}
\resizebox{\linewidth}{!}{%
\begin{tabular}{lc|cccccccc|cc}
\toprule
       &         & \multicolumn{8}{c|}{Short dependency QA}                                            & \multicolumn{2}{c}{Cloze} \\
\multicolumn{1}{p{2em}}{\textbf{Models}} & 
\multicolumn{1}{p{3em}|}{\textbf{Context}} & 
\multicolumn{1}{p{2em}}{\textbf{Bleu1}} & 
\multicolumn{1}{p{2em}}{\textbf{Bleu4}} &
\multicolumn{1}{p{3em}}{\textbf{Rouge1}} &
\multicolumn{1}{p{3em}}{\textbf{Rouge4}} & 
\multicolumn{1}{p{3em}}{\textbf{RougeL}} &    
\multicolumn{1}{p{6em}}{\textbf{Meteor\_score}} &
\multicolumn{1}{p{6em}}{\textbf{Bert\_score}} & 
\multicolumn{1}{p{6em}|}{\textbf{GPT4\_score}} &   
\multicolumn{1}{p{6em}}{\textbf{Exact Match}} & 
\multicolumn{1}{p{6em}}{\textbf{Partial Match}} \\ 
    \midrule
GPT4-32k         & \cellcolor{lightgray!70}32k              & 24.61 & 11.14      & 61.80 & 50.73 & 60.75 & 32.94 & 78.72 & \textbf{71.52} & \textbf{70.5} & \textbf{80.81}\\
GPT4-8k          & \cellcolor{lightgray!30}8K               & 27.35 & 14.38      & \textbf{67.59} & \textbf{56.01} & \textbf{65.77} & \textbf{38.56} & \textbf{87.93} & 53.99 & 66.03 & 76.62\\
GPT3.5-turbo-16k & \cellcolor{lightgray!50}16K              & 22.67 & 9.62       & 62.56 & 48.63 & 60.66 & 32.58 & 87.04 & 66.82 & 54.64 & 63.42 \\
LlamaIndex       & \textbackslash{} & \textbf{33.37} & \textbf{21.43}     & 58.82 & 42.93 & 57.08 & 37.17 & 86.58 & 59.61 & 58.95 & 66.86\\
ChatGLM2-6B       & \cellcolor{lightgray!70}32k              & 14.29 & 6.07       & 20.50  & 13.16 & 20.36 & 13.08 & 87.28 & 23.65 & 0.05 & 0.98 \\
LongLLaMa-3B   & \cellcolor{lightgray}256k             & 1.37  & 0.26       & 26.97 & 11.02 & 26.10  & 11.34 & 71.65 & 13.75 &  - & 2.13 \\
RWKV-4-14B-pile         & \cellcolor{lightgray!30}8k               & 0.80   & 0.04       & 21.7  & 6.39  & 20.64 & 9.41  & 70.42 & 8.93 & - & - \\
LLaMA2-7B-32K        & \cellcolor{lightgray!70}32k              & 0.18  & 7.25*e-308 & 1.86  & 0.00     & 1.86  & 1.52  & 61.53 & 3.18 & - & 0.58 \\
\bottomrule
\end{tabular}
}
\end{table}

\begin{table}[t!]
\caption{Performance of the long dependency tasks}
\label{tab:long_dependency_tasks_performence}
\resizebox{\linewidth}{!}{%
\begin{tabular}{lccccccccc}
\toprule
\multicolumn{1}{p{2em}}{\textbf{Models}} & 
\multicolumn{1}{p{3em}}{\textbf{Context}} & 
\multicolumn{1}{p{2em}}{\textbf{Bleu1}} & 
\multicolumn{1}{p{2em}}{\textbf{Bleu4}} &
\multicolumn{1}{p{3em}}{\textbf{Rouge1}} &
\multicolumn{1}{p{3em}}{\textbf{Rouge4}} & 
\multicolumn{1}{p{3em}}{\textbf{RougeL}} &    
\multicolumn{1}{p{6em}}{\textbf{Meteor\_score}} &
\multicolumn{1}{p{6em}}{\textbf{Bert\_score}} & 
\multicolumn{1}{p{6em}}{\textbf{GPT4\_score}} \\   
\midrule
arXiv paper summarization \\
\midrule
GPT4-32k                  & \cellcolor{lightgray!70}32k              & 24.50 & 0.73      & 27.15 & 7.10     & 24.25 & 19.03 & 84.04 & 82.84 \\
GPT4-8k                   & \cellcolor{lightgray!30}8k               & \textbf{29.02} & \textbf{2.09}      & \textbf{32.08} & \textbf{11.11}    & 28.85 & \textbf{22.64} & \textbf{84.92} & 85.42 \\
GPT3.5-turbo-16k          & \cellcolor{lightgray!50}16k              & 28.70 & 1.59      & 32.04 & 10.69    & \textbf{28.89} & 22.34 & 84.82 & \textbf{86.84} \\
LlamaIndex               & \textbackslash{} & 22.53 & 0.63      & 26.28 & 6.97     & 23.73 & 21.07 & 83.09 & 76.35 \\
ChatGLM2-6B                & \cellcolor{lightgray!70}32k              & 0.04  & 1.60e-310 & 5.97  & 8.43e-05 & 5.82  & 6.40  & 73.25 & 13.23 \\
LongLLaMa-3B            & \cellcolor{lightgray}256k             & 4.24  & 9.32e-309 & 4.10  & 0.52     & 3.86  & 3.82  & 73.41 & 12.28 \\
RWKV-4-14B-pile                  & \cellcolor{lightgray!30}8k               & 6.28  & 4.58e-05  & 6.45  & 0.74     & 6.01  & 6.00  & 75.28 & 7.02  \\
LLaMA2-7B-32K                 & \cellcolor{lightgray!80}32k              & 0.03  & 4.66e-310 & 0.12  & 0.00     & 0.12  & 0.67  & 71.21 & 7.60   \\
\midrule
Long dependency QA\\  
\midrule
GPT4-32k                  & \cellcolor{lightgray!70}32k              & 8.55  & 1.40      & \textbf{25.59} & 6.36     & \textbf{24.04} & \textbf{11.13} & 80.16 & \textbf{54.09} \\
GPT4-8k                  & \cellcolor{lightgray!30}8k               & \textbf{8.94}  & 1.01      & 23.45 & 6.57     & 21.69 & 10.18 & 85.36 & 42.12 \\
GPT3.5-turbo-16k          & \cellcolor{lightgray!50}16k              & 6.92  & \textbf{1.81}     & 25.02 & 6.68     & 23.63 & 10.40 & 83.79 & 45.04 \\
LlamaIndex               & \textbackslash{} & 7.76  & 1.24      & 23.62 & \textbf{7.10}     & 22.30 & 10.47 & 83.87 & 37.63 \\
ChatGLM2-6B                & \cellcolor{lightgray!70}32k              & 5.55  & 0.11      & 9.41  & 1.93     & 8.69  & 4.39  & \textbf{85.78} & 11.50 \\
LongLLaMa-3B            & \cellcolor{lightgray}256k             & 1.04  & 3.12e-307 & 2.96  & 0.03     & 2.71  & 1.66  & 78.60 & 6.48  \\
RWKV-4-14B-pile                  & \cellcolor{lightgray!30}8k               & 0.71  & 9.52e-307 & 18.54 & 1.55     & 17.69 & 3.45  & 71.36 & 5.33  \\
LLaMA2-7B-32K                 & \cellcolor{lightgray!70}32k              & 0.08  & 2.44e-308 & 2.05  & 0.00        & 2.05  & 0.46  & 50.28 & 4.18  \\
\bottomrule
\end{tabular}}
\end{table}

\subsection{Results}

\cref{fig:performance} shows an overall performance comparison of different models on different tasks. 
The first radar plot shows the original accuracy evaluated by GPT4-8k (except cloze) and the partial match result (for cloze) over different tasks. For better visualization, we scale the scores of all models on each task to [40, 100] in the second radar plot and the histogram, so that the best model on each task has a score of 100 and the worst model has a score of 40. From the charts, GPT4-32k demonstrates its impressive overall performance across all tasks (with highest scores on all tasks except summarization). In comparison, open-source models show a significant performance gap to commercial models on our benchmark. From the first radar chart, we can find that among the 7 major tasks, short QA, cloze and summarization are more effectively addressed by LLMs, while \textbf{real long dependency QA tasks are far from being solved, where even GPT4-32k hardly achieves over 40\% accuracy}. The empirical results demonstrate that even the most successful commercial model still cannot effectively address those really challenging long dependency tasks, leaving large room for improvement. Detailed evaluation results and further analysis can be found in the following sections.

\subsubsection{Main results on long and short dependency tasks}

\paragraph{Results on short dependency tasks~}
\cref{tab:short_term_qa_performence} presents the performance (\%) of all the baselines on \dataset in short dependency tasks.
Notably, GPT4-32k attains the highest accuracy according to the GPT4 evaluator's perspective. GPT4-8k, GPT3.5-turbo-16k, and the retrieval-based LlamaIndex closely follow, demonstrating competitive performance levels. Surprisingly, GPT4-8k exhibits the most robust overall performance in terms of automatic evaluation metrics. It's worth mentioning that GPT4-32k, due to its tendency to generate longer outputs, faces penalties from these automatic metrics. This discrepancy among different metrics highlights the need for improved evaluation methods. Furthermore, in the context of cloze tasks, GPT4-32k excels again when equipped with a longer context window. In \cref{fig:output}, the exact match results in cloze tasks are displayed for varying source segment lengths. The results show that as the segment length increases, model performance gradually decreases, underscoring the increasing difficulty of effectively filling in the masked entities with longer source text.

\paragraph{Results on long dependency tasks~} 
\cref{tab:long_dependency_tasks_performence} shows the aggregated results on long dependency tasks. Firstly, we can observe that summarization can be well addressed by commercial models, with GPT-4 evaluation accuracy of over 80\%. However, the various types of long dependency QAs in our benchmark apparently pose substantial challenges for current LLMs. Both open-source and commercial models experience a significant performance decline. We will analyze model performance on individual types of QAs in \cref{sec:diverse}.
It is validated that longer context window size (thus less information loss due to truncation) indeed helps in long context tasks by comparing GPT4-32k with GPT4-8k. GPT4-8k has a much lower accuracy by answering ``The text does not provide information on ..." in many cases. Open-sourced models fall far below the average of commercial models, among which LLaMA2-7B-32K and RWKV-4-14B-pile display almost zero performance. By employing context scaling techniques like positional interpolation, RNN and fine-tuning on longer texts, current LLMs can be equipped with much longer context windows than their default limits. Nevertheless, our results show that there is still a huge discrepancy between merely increasing the context window size and really understanding the long context. 
The poor performance on long dependency QAs suggests that we may need to revisit LLMs' long context understanding ability in more challenging tasks other than some simple ones like summarization and retrieval, as they are unable to test whether LLMs understand the inter-dependency in long texts.  

\begin{table}[t!]
\caption{Impact of input length on long dependency tasks}
\label{tab:impact_of_context_length}
\resizebox{\linewidth}{!}{%
\begin{tabular}{lccccccccc}
\toprule
\multicolumn{1}{p{2em}}{\textbf{Models}} & 
\multicolumn{1}{p{3em}}{\textbf{Context}} & 
\multicolumn{1}{p{3em}}{\textbf{Bleu1}} & 
\multicolumn{1}{p{3em}}{\textbf{Bleu4}} &
\multicolumn{1}{p{4em}}{\textbf{Rouge1}} &
\multicolumn{1}{p{4em}}{\textbf{Rouge4}} & 
\multicolumn{1}{p{4em}}{\textbf{RougeL}} &    
\multicolumn{1}{p{6em}}{\textbf{Meteor\_score}} &
\multicolumn{1}{p{6em}}{\textbf{Bert\_score}} & 
\multicolumn{1}{p{6em}}{\textbf{GPT4\_score}} \\   
\midrule
\multicolumn{8}{l}{arXiv paper summarization}  \\
\midrule
GPT4-32k           & \cellcolor{lightgray!70}32k & 24.50 & 0.73 & 27.15 & 7.10  & 24.25 & 19.03 & 84.04 & 82.84 \\
GPT4-32k           & \cellcolor{lightgray!60}24k & 25.57 & 0.81 & 27.61 & 7.53 & 24.73  & 19.86 &  84.07  & 83.15   \\
GPT4-32k           & \cellcolor{lightgray!50}16k & 24.80 & 0.70 & 27.29 & 7.26  & 24.28 & 19.12 & 84.11 & 82.82 \\
GPT4-32k           & \cellcolor{lightgray!30}8k  & 26.26 &  \textbf{9.35} & 27.83 & 7.67  & 24.74 & 20.08 & 84.10 & 82.75 \\
GPT4-8k               & \cellcolor{lightgray!30}8k  & \textbf{29.02} & 2.09 &  \textbf{32.08} &  \textbf{11.11} &  \textbf{28.85} &  \textbf{22.64} &  \textbf{84.92} &  \textbf{85.42} \\
\midrule
\multicolumn{8}{l}{Long dependency QA}  \\
\midrule
GPT4-32k           & \cellcolor{lightgray!70}32k & 7.64  & 1.24 & 15.53 & 4.46  & 14.60 & 11.12 & 86.07 & \textbf{54.65} \\
GPT4-32k           & \cellcolor{lightgray!60}24k & 8.23  & 1.66 & 14.92 & 4.12  & 13.90 & 10.60 & 86.16 & 50.61 \\
GPT4-32k           & \cellcolor{lightgray!50}16k & 8.57  & 1.35 & 16.21 & 4.30  & 14.90 & \textbf{11.91} & \textbf{86.36} & 47.55 \\
GPT4-32k           & \cellcolor{lightgray!30}8k  & 7.46  & \textbf{1.77} & 13.75 & 5.08  & 12.89 & 10.01 & 85.77 & 38.34 \\
GPT4-8k               & \cellcolor{lightgray!30}8k  & \textbf{8.94}  & 1.01 & \textbf{23.45} & \textbf{6.57}  & \textbf{21.69} & 10.18 & 85.36 & 42.12 \\
\bottomrule
\end{tabular}
}
\end{table}

\begin{table}[t!]
\caption{Performance of the long dependency QA with LlamaIndex}
\label{tab:long_dependency_tasks_performence_llamaindex}
\resizebox{\linewidth}{!}{%
\begin{tabular}{lccccccccc}
\toprule
\multicolumn{1}{p{2em}}{\textbf{Models}} & 
\multicolumn{1}{p{3em}}{\textbf{Context}} & 
\multicolumn{1}{p{2em}}{\textbf{Bleu1}} & 
\multicolumn{1}{p{2em}}{\textbf{Bleu4}} &
\multicolumn{1}{p{3em}}{\textbf{Rouge1}} &
\multicolumn{1}{p{3em}}{\textbf{Rouge4}} & 
\multicolumn{1}{p{3em}}{\textbf{RougeL}} &    
\multicolumn{1}{p{6em}}{\textbf{Meteor\_score}} &
\multicolumn{1}{p{6em}}{\textbf{Bert\_score}} & 
\multicolumn{1}{p{6em}}{\textbf{GPT4\_score}} \\    
\midrule
GPT4-32k        & \cellcolor{lightgray!70}32k       & 6.08  & 1.31      & 10.27 & 3.39     & 9.52 & 8.54 & 85.27 & 28.25 \\
GPT4-8k         & \cellcolor{lightgray!30}8k        & 6.62  & \textbf{1.50}     & \textbf{11.95} & \textbf{3.80}     & \textbf{10.99} & 9.02 & 85.51 & 26.34 \\
GPT3.5-turbo-16k          & \cellcolor{lightgray!50}16k    & 6.50  & 0.92     & 10.93 & 3.56     & 9.86 & 8.65 & \textbf{85.63} & \textbf{33.24} \\
Default               & \textbackslash{} & \textbf{7.02}  & 1.24      & 11.60 & 3.75     & 10.57 & \textbf{9.37} & 85.61 & 33.16 \\
ChatGLM2-6B-32k               & \cellcolor{lightgray!70}32k     & 0.15  & 2.82e-310      & 2.23  & 0.05     & 2.23  & 0.74  & 83.40 & 7.73 \\
LongLLaMa-3B            & \cellcolor{lightgray}256k    & 1.04  & 4.22e-311 & 2.27  & 0.00    & 2.23  & 2.23  & 82.18 & 5.33  \\
RWKV-4-14B-pile                  & \cellcolor{lightgray!30}8k    & 2.65  & 8.09e-307 & 4.08 & 0.33     & 3.65 & 3.92  & 80.74 & 2.43  \\
LLaMA2-7B-32K                 & \cellcolor{lightgray!70}32k    & 0.43  & 1.29e-307 & 5.85  & 0.00   & 5.85  & 1.04  & 81.38 & 6.76  \\
\bottomrule
\end{tabular}}
\end{table}

\subsubsection{Deep dive into long context understanding capabilities}\label{sec:diverse}
In this section, we analyze different factors affecting the long context understanding abilities of LLMs, and dive into individual types of long dependency QAs to check LLMs' limitations.

\paragraph{Impact of varying input length~} In \cref{tab:impact_of_context_length}, we study the impact of varying lengths of inputs on long dependency tasks with GPT4 models. We find that expanding input length hardly helps in paper summarization while it substantially enhances the model's performance on long dependency QAs. The difference can be attributed to the inherent nature of the arXiv paper. It has both the introduction and conclusion sections located at the beginning and in the end respectively, which already contain the major sketch of the paper. Meanwhile, in our expectation, longer input promotes the performance of long dependency QAs by introducing less information loss. 

\paragraph{Retrieval Based Techniques~} 
To evaluate the effectiveness of retrieval techniques for long-context dependency questions, we undertook an extensive series of experiments on our long dependency QA tasks by replacing the base LLM model in LlamaIndex with different baseline LLMs. In these experiments, we utilized the open-source embedding all-mpnet-base-v2 ~\citep{song2020mpnet}. When compared to the default embedding, text-embedding-ada-002 (https://openai.com/blog/new-and-improved-embedding-model), there was a noticeable performance decline. Nonetheless, this disparity did not hinder our conclusions.
 From \cref{tab:long_dependency_tasks_performence} and \cref{tab:long_dependency_tasks_performence_llamaindex}, our research findings reveal that the incorporation of retrieval techniques does not generally enhance the performance of long dependency QA tasks. There is a conspicuous performance decline, particularly evident for models like GPT4-8k and GPT4-32k. It can be attributed to the tendency of GPT models to produce longer outputs, sometimes including hallucinatory information, when the retrieved segments lack sufficient context. The phenomenon highlights the intricacy of our dataset, where a series of long dependency understanding and modeling capabilities such as comprehension and multi-hop reasoning are essentially needed. Relying solely on retrieval mechanisms might be insufficient in recalling the necessary information and further generating the final answer, resulting in a marked performance decline. However, we did observe an minor improvement in the BERT score for open-source models. This improvement in fluency can be attributed to the considerably shorter length of the retrieved segments used as inputs, in contrast to the entirety of the document.
 

\paragraph{Individual results on different types of long dependency QAs~} 
Previous research mostly focuses on presenting aggregated results for long dependency QA tasks across various question types. Differently, in this study, our objective is to delve into the performance of models in individual tasks that demand diverse capabilities, including reading comprehension, information retrieval, computation, and reasoning. In this regard, we employed GPT4 as the evaluator, and the accuracy results are available in \cref{tab:long_term_qa_evaluated_by_gpt4}. Across the four tasks examined, LLMs generally exhibit strong performance in comprehension, reasoning, and multiple information retrieval, but fall short in tasks related to timeline reordering and computation. Furthermore, we observed that the way questions are framed has a significant impact on LLMs' performance. Yes-no questions and multiple-choice questions tend to be easier for LLMs to answer, particularly when the search space is limited, as opposed to open-ended questions within unstructured text.

\begin{table}[t]
\vspace{-0.5cm}
\caption{Individual results on four types of long dependency QAs evaluated by GPT4}
\label{tab:long_term_qa_evaluated_by_gpt4}
\resizebox{\linewidth}{!}{%
\begin{tabular}{lccccc}
\toprule
\multicolumn{1}{p{3em}}{\textbf{Models}} & 
\multicolumn{1}{p{3em}}{\textbf{Context}} & 
\multicolumn{1}{p{12em}}{\textbf{\makecell[c]{Information retrieval}}} &    
\multicolumn{1}{p{12em}}{\textbf{\makecell[c]{Timeline reorder}}} &
\multicolumn{1}{p{6em}}{\textbf{Computation}} & 
\multicolumn{1}{p{18em}}{\textbf{\makecell[c]{Comprehension and reasoning}}} \\   
    \midrule
GPT4-32k         & \cellcolor{lightgray!70}32k              & \textbf{43.60} & \textbf{64.43} & \textbf{37.36} & \textbf{61.26} \\
GPT4-8k          & \cellcolor{lightgray!30}8K               &31.89  &61.36 & 22.54 &45.78  \\
GPT3.5-turbo-16k & \cellcolor{lightgray!50}16K              & 36.86& 55.73 & 24.73 &51.09  \\
LlamaIndex       & \textbackslash{} & 27.60 & 47.83 & 19.78 &  43.83\\
ChatGLM2-6B-32k       & \cellcolor{lightgray!70}32k  & 12.47 &14.17  &5.43  &11.08  \\
LongLLaMa-3B   & \cellcolor{lightgray}256k     & 3.82 & 6.48 & 5.15 & 10.17 \\
RWKV-4-14B-pile         & \cellcolor{lightgray!30}8k  &4.67  & 5.19 & 4.40 & 7.13\\
LLaMA2-7B-32K        & \cellcolor{lightgray!70}32k    &  3.01 & 1.61  & 1.12 & 6.85  \\
\bottomrule
\end{tabular}
}
\end{table}

\begin{table}[t!]
\caption{Performance of Timeline Reorder}
\label{tab:timeline_reorder_performence}
\resizebox{\linewidth}{!}{%
\begin{tabular}{lccccccccc}
\toprule
\multicolumn{1}{p{2em}}{\textbf{Models}} & 
\multicolumn{1}{p{2em}}{\textbf{LSD}} & 
\multicolumn{1}{p{2em}}{\textbf{LMD}} &    
\multicolumn{1}{p{2em}}{\textbf{SD}} &
\multicolumn{1}{p{2em}}{\textbf{SDD}} & 
\multicolumn{1}{p{3em}}{\textbf{LSD-S}} & 
\multicolumn{1}{p{4em}}{\textbf{LMD-S}} &    
\multicolumn{1}{p{3em}}{\textbf{SD-S}} &
\multicolumn{1}{p{3em}}{\textbf{SDD-S}} & 
\multicolumn{1}{p{10em}}{\textbf{Non-standard(\%)}} \\   
    \midrule
GPT4-32k         & \textbf{1.04} & \textbf{0.57} & \textbf{0.93} & \textbf{1.12} & 1.21 & \textbf{0.82} & 1.41 & 1.60 & 52.80 \\
GPT4-8k          & 1.24 & 0.64 & 1.04  & 1.281 & 1.43 & 0.92  & 1.51 & 1.74 & 49.31 \\
LlamaIndex       & 1.55 & 0.78 & 1.19  & 1.551 & 1.95 & 1.08  & 1.65 & 2.09 & \textbf{39.72} \\
GPT3.5-turbo-16k & 3.58 & 1.43  & 2.17  & 2.916 & \textbf{1.05} & 0.86  & 1.24 & \textbf{1.26} & 77.21 \\
LongLLaMa-3B   & 4.18 & 1.71  & 2.59  & 3.30 & 1.90 & 1.12  & 1.80 & 2.07 & 92.92 \\
ChatGLM2-6B-32k          & 4.31 & 1.74  & 2.63  & 3.37 & 1.83 & 1.17  & 1.50 & 2.00 & 99.07 \\
RWKV-4-14B-pile         & 4.33 & 1.75  & 2.64  & 3.38 & 1.90 & 0.97 & \textbf{1.00} & 1.75 & 98.13 \\
LLaMA2-7B-32K        & 4.33 & 1.75  & 2.64  & 3.38 & 2.50 & 1.17  & 1.33 & 2.33 & 98.60 \\ 
\bottomrule
\end{tabular}
}
\end{table}

\begin{figure}[b!]
    \begin{minipage}[t]{0.48\textwidth}
    \centering
    \includegraphics[width=\linewidth]{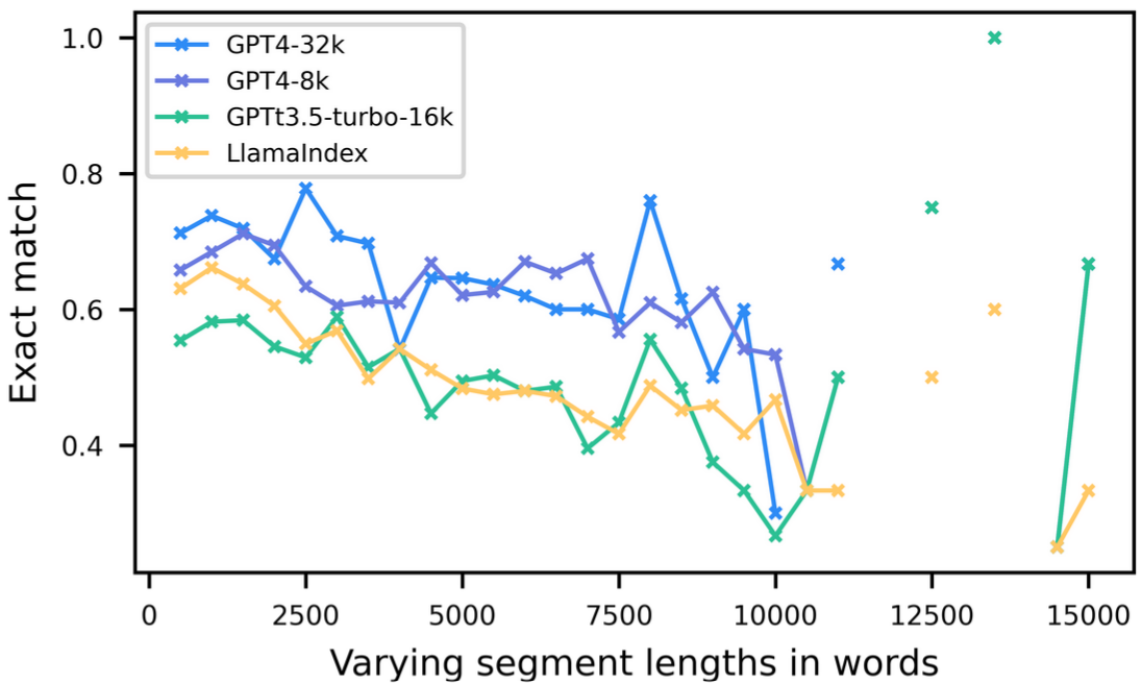}
    \caption{Performance of varying segments}
    \label{fig:output}\hfill
  \end{minipage}%
    \begin{minipage}[t]{0.46\textwidth}
    \centering
    \includegraphics[width=\linewidth]{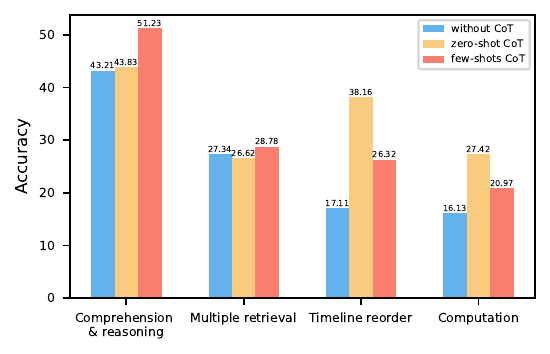}
    \caption{Long dependency QA tasks w/o CoT}
    \label{fig:cot}
  \end{minipage}
\end{figure}

\paragraph{Results on long dependency QAs with/without CoT~} 
To bolster the long-context capabilities of LLMs, we conducted additional experiments designed to unlock their potential using the Chain of Thoughts~(CoT) framework~\citep{kojima2023large}. We selected LlamaIndex as a representative model, given its impressive performance in both short and long dependency question-answering tasks, alongside strong commercial models such as GPT4. A manual evaluation was carried out on a subset comprising one-third of instances from each task category within long dependency QA. We initiated the LLM with a zero-shot CoT approach, employing prompts such as ``Let's think step by step," and furnished a few-shot setup with detailed rationales and standard output formats~\cite{wei2023chainofthought} to facilitate responses to long dependency questions. As depicted in \cref{fig:cot}, the zero-shot CoT approach had minimal impact on accuracy in comprehension and reasoning, as well as multiple retrieval tasks, but yielded a substantial 20\% and 10\% absolute accuracy increase in timeline reorder and computation. Interestingly, the few-shot CoT approach benefits the first two types but surprisingly leads to a decline in performance in the latter two types compared with zero-shot.
We hypothesize the reason is that the evidence and rationales in few-shot examples cannot be generalized to other questions, and including them might on the contrary give wrong guidance to the model. 

\paragraph{Automatic evaluation on timeline reorder} 
In order to evaluate the performance of time reorder task outputs, it is essential to address discrepancies arising from the diverse formats produced by various models. Typically, these outputs comprise conventional numerical sequences, but errors in non-standard formats when evaluation necessitate preprocessing for accurate assessment. A proposed approach involves converting the serial numbers in the candidate answers from their original question into Roman numbers (\ie, I, II, $\cdots$), thereby enhancing discrimination through regular expression matching. Four key metrics, namely, LSD (location square deviation), LMD (location mean deviation), SD (swap deviation), and SDD (swap distance deviation), are employed to measure the similarity of numeric sequences, refer to Appendix~\ref{appendix:reordermetrics} for metric details. Smaller deviations indicate a higher degree of resemblance between the sequences. Any outputs that are empty, possess unequal lengths, or contain extra elements are categorized as non-standard. The maximum deviation between the provided ground truth and all corresponding candidate answers is computed as the worst score for evaluation purposes. 

The percentage of non-standard outputs for each model and corresponding performances can be found in \cref{tab:timeline_reorder_performence}.
As seen, it is evident that except for GPT4, which demonstrates a remarkable degree of adherence and alignment following Reinforcement Learning from Human Feedback (RLHF)~\citep{lee2023rlaif}, most open-sourced models struggle to generate texts in the correct format with less than 10\%. However, this issue can be mitigated in significantly large models through the utilization of few-shot 
prompts and mandatory instructions. This phenomenon results in performance 
penalties when assessed using automated metrics. Consequently, to ascertain the
genuine capacity of LLMs in this task, we calculate the four metrics exclusively for outputs in standard format (``-S''). 

\begin{wrapfigure}{r}{.6\textwidth}
    \centering
    \includegraphics[width=\linewidth]{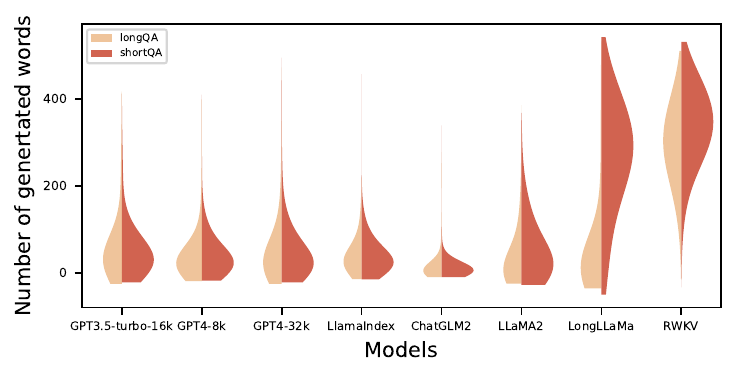}
    \caption{Output distributions on QA tasks}
    \label{fig:generated_words}
    \vspace{-.2in}
\end{wrapfigure}

\paragraph{Dispcrepancy in generated outputs of models} 
Distributions of generated outputs of various models are depicted in \cref{fig:generated_words}. 
It is noteworthy that well-behaved models consistently produce shorter responses, averaging around 50 words, irrespective of the question type, particularly in short-term question answering scenarios. In contrast, models fine-tuned with longer textual inputs, such as LLaMA2-7B-32K, RWKV-4-14B-pile, and LongLLaMa-3B, tend to yield significantly lengthier responses, even when a maximum generation constraint of 500 tokens is enforced. An interesting deviation is observed in LongLLaMa-3B, which demonstrates variability in response lengths across both tasks. This behavior may stem from challenges in comprehending and addressing exceedingly complex long question-answering tasks. Consequently, the model appears to prioritize extracting a maximum number of pertinent contexts from its memory to generate sufficiently extensive responses that are deemed acceptable and rational.

Moreover, upon closer examination of model outputs, a significant disparity in generation quality is observed across various LLMs and task types, indicating a non-specific issue. Notably, commercial models like GPT4, GPT3.5, and LlamaIndex consistently generate outputs that exhibit a higher degree of human-likeness, completeness, and logical coherence within a structured format. These models consistently deliver contextually relevant, query-based responses. In contrast, open-sourced models, such as ChatGLM2-6B-32k, tend to offer shorter answers, occasionally confined to numeric responses. In cases where a definite answer is lacking, ChatGLM2-6B-32k compensates by retrieving relevant contextual information. However, the RNN-based model RWKV-4-14B-pile often generates duplicated responses or resorts to repeating the given questions to reach the maximum token length, sometimes resorting to code generation to address issues related to its training data. The performance of the LLaMA2-7B-32K model is notably worse, as it sporadically produces irrelevant or nonsensical text, along with the inclusion of special symbols when it fails to provide meaningful answers. More examples of outputs from different models can be seen in Appendix~\ref{sec:example_output}.

\paragraph{Probable explanations for long QA bad cases}

\begin{table}[b]
\vspace{-0.2in}
\caption{Bad cases study on the long dependency QA}
\label{tab:failure_case_analysis}
\resizebox{\textwidth}{!}{  
\begin{tabular}{lccccccc}
\toprule
\makecell[c]{Long  QA  Tasks} & Hallucination$^*$ & \makecell{Redundant \\retrieval$^\dagger$} & \makecell{Insufficient \\retrieval$^\star$}     & \makecell{Irrelevant \\answer$^\diamond$}  &  \makecell{No relevant \\context$^\land$}  & \makecell{Wrong/No \\reasoning$^\times$}   &  Others    \\
\midrule
\makecell[c]{Computation}  & 31.11 & 24.44 & 15.56 & 0.00     & 20.00 & 0.00     & 8.88 \\
\makecell[c]{Multiple information retrieval}  & 14.71 & 31.37 & 28.43 & 13.73 & 13.73 & 0.00     & 7.84 \\
\makecell[c]{Comprehension and reasoning} & 14.29 & 10.99 & 21.98 & 18.68 & 16.48 & 10.99 & 6.59 \\
\bottomrule
\multicolumn{8}{l}{\footnotesize $^*$ Evidence of predictions is not shown up in the original inputs and generated by LLM itself from nowhere.} \\
\multicolumn{8}{l}{\footnotesize $^\dagger$ Apart from the right evidence, irrelevant evidence is also redundantly retrieved.} \\
\multicolumn{8}{l}{\footnotesize $^\star$ Not all of the essential evidence to answer the question is retrieved.} \\
\multicolumn{8}{l}{\footnotesize $^\diamond$ Evidence of predictions generated have no or minor correlation with the question.} \\
\multicolumn{8}{l}{\footnotesize $^\land$ No relevant context in LLM's memory and refuse to answer the question.} \\
\multicolumn{8}{l}{\footnotesize $^\times$ Fail to retrieve evidence that needs further reasoning other than directly extracting from the inputs.} \\
\end{tabular}
}
\end{table}

To investigate whether the models have effectively memorized and comprehended lengthy contextual information, we conducted a comprehensive manual analysis of the underlying causes of failures in each long question-answering task. The rationale behind CoT analysis aided in understanding how models decompose and tackle challenges associated with extended dependency-based QA. Our observations reveal that LLMs struggle with these tasks primarily due to their inability to extract precise information and a propensity to generate responses that lack factual accuracy. Constraints imposed by the inherent context window limitations, coupled with information loss resulting from the optimized Transformer and position encoding, contribute to their struggles in memorizing the original extensive contexts. In most cases, models attempt to compensate by retrieving and integrating the most pertinent evidence, even if it results in redundant answers. However, they also acknowledge their insufficient context and, at times, abstain from providing responses rather than resorting to nonsensical answers. Furthermore, addressing these challenges necessitates enhanced comprehension and reasoning abilities, particularly when answers are not clearly evident across multiple pieces of evidence scattered throughout the raw texts. The insights derived from our benchmark analysis offer a scientific foundation and pave the way for promising research directions aimed at augmenting LLM capabilities for handling long contextual inputs. These findings underscore the need for further progress in comprehension, computation, and reasoning tasks using our dataset to effectively enhance LLMs' capacity to understand extended dependency contexts.

\section{Conclusion}

This paper introduces a novel benchmark, \dataset, designed to facilitate the assessment of long-context comprehension by LLMs. \dataset addresses the deficiencies present in previous datasets by offering considerably longer text passages, utilizing relatively new documents after 2022, incorporating multi-source materials from various categories, and notably featuring meticulously designed and annotated tasks with diverse contextual dependencies. Our extensive evaluations unveil substantial limitations in the capacity of existing LLMs to understand and reason about the intricate interdependencies present in lengthy texts, even when provided with considerably extended context windows. Furthermore, a notable disparity is observed between commercial and open-source models, with both exhibiting challenges in long dependency tasks as per our benchmark assessments. The outcomes underscore the utility of our dataset as a valuable reference for evaluating long-context comprehension and present avenues for potential enhancements in LLM performance.

\clearpage
\bibliography{iclr2024_conference}
\bibliographystyle{iclr2024_conference}

\clearpage
\appendix

\section{More details of our dataset}
\begin{figure}[htbp]
  \begin{minipage}[t!]{0.5\textwidth}
    \centering
    \includegraphics[width=\linewidth]{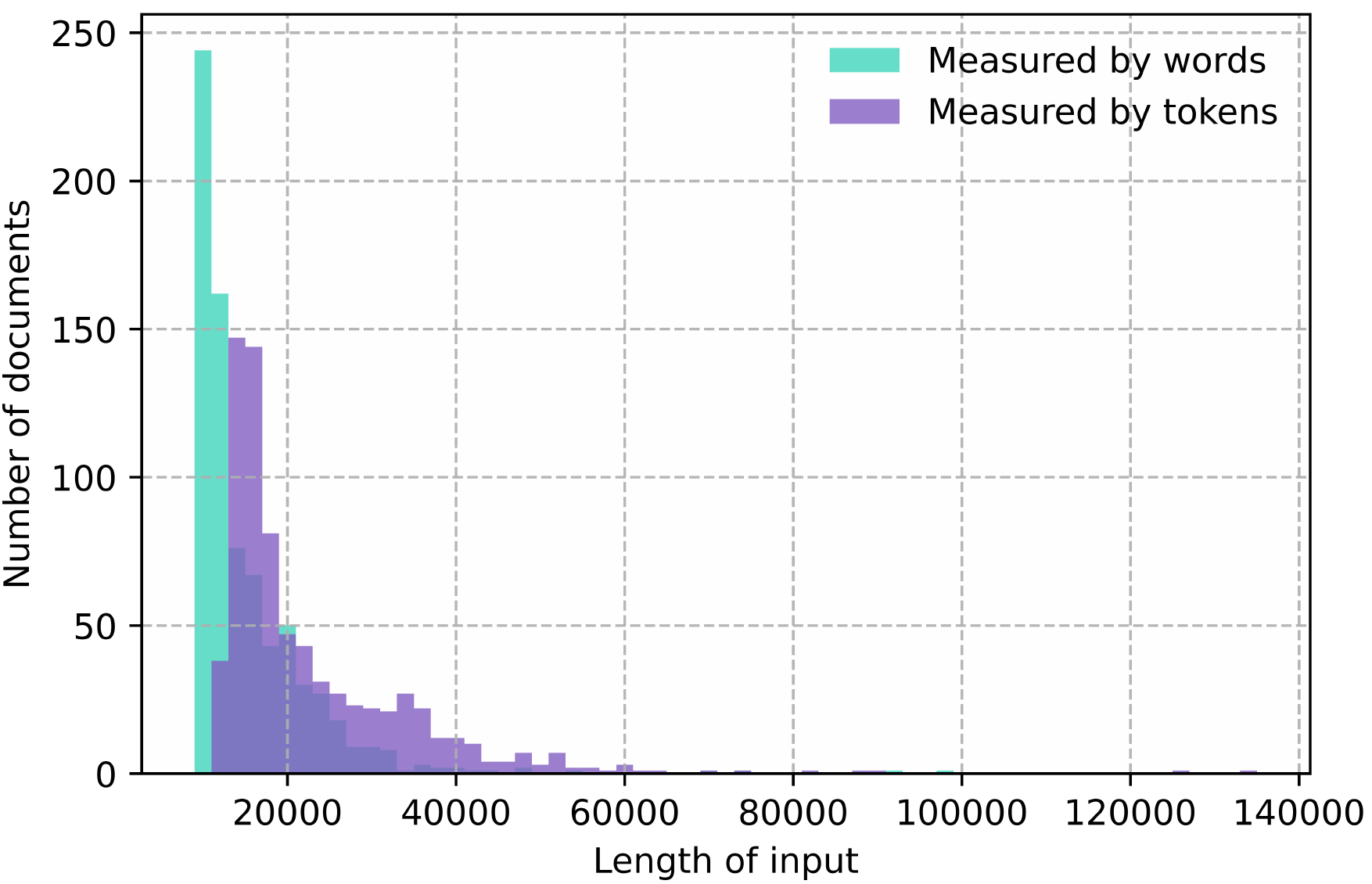}
    \caption{Input document length distributions}
    \label{fig:inputdis}
  \end{minipage}%
  \begin{minipage}[t!]{0.49\textwidth}
    \centering
    \includegraphics[width=\linewidth]{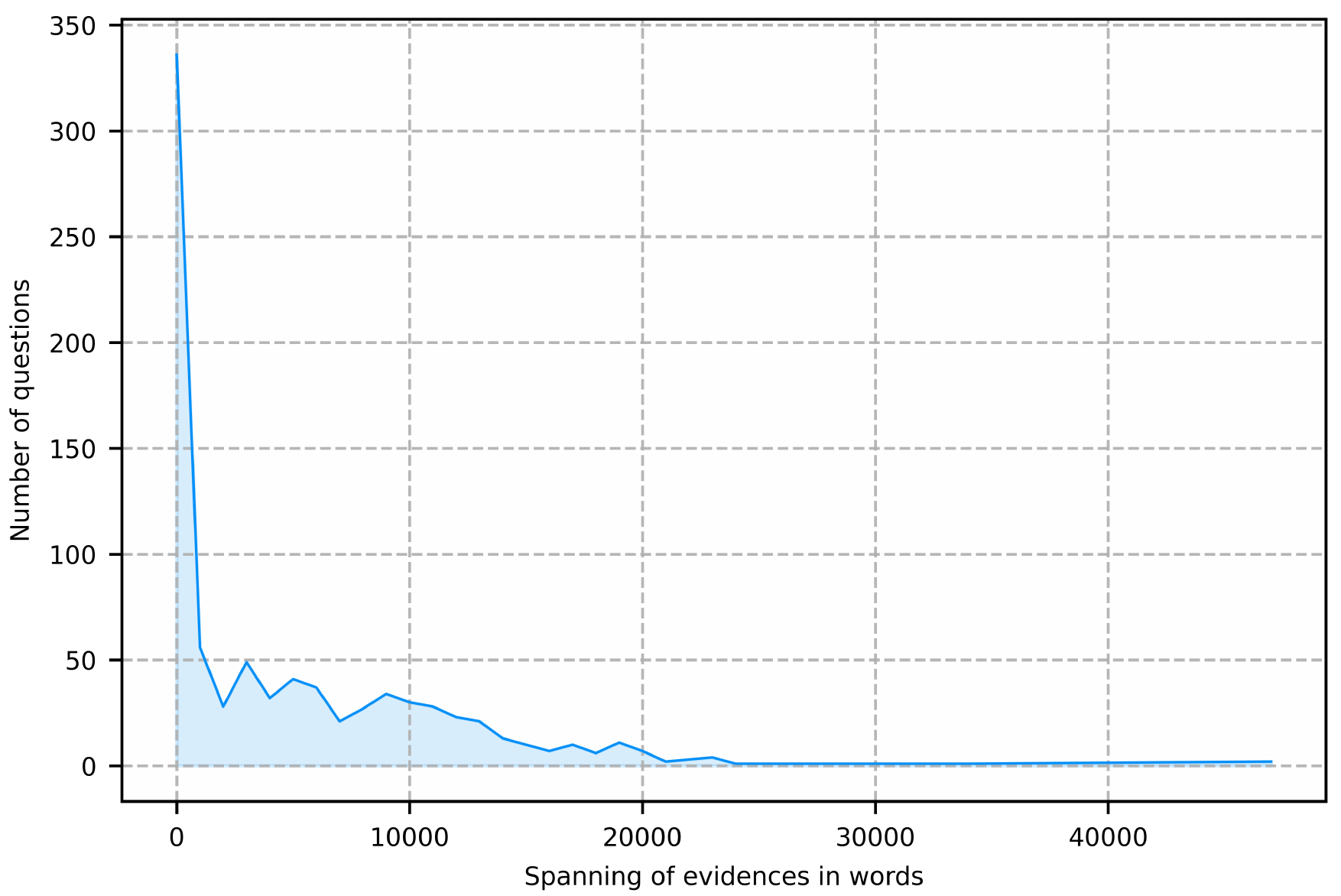}
    \caption{Evidence dependency distributions}
    \label{fig:evddis}
  \end{minipage}
\end{figure}

\begin{figure}[htbp]   
  \centering             
  {
      \label{fig:subfig1}\includegraphics[width=0.25\textwidth,height=0.25\textwidth]{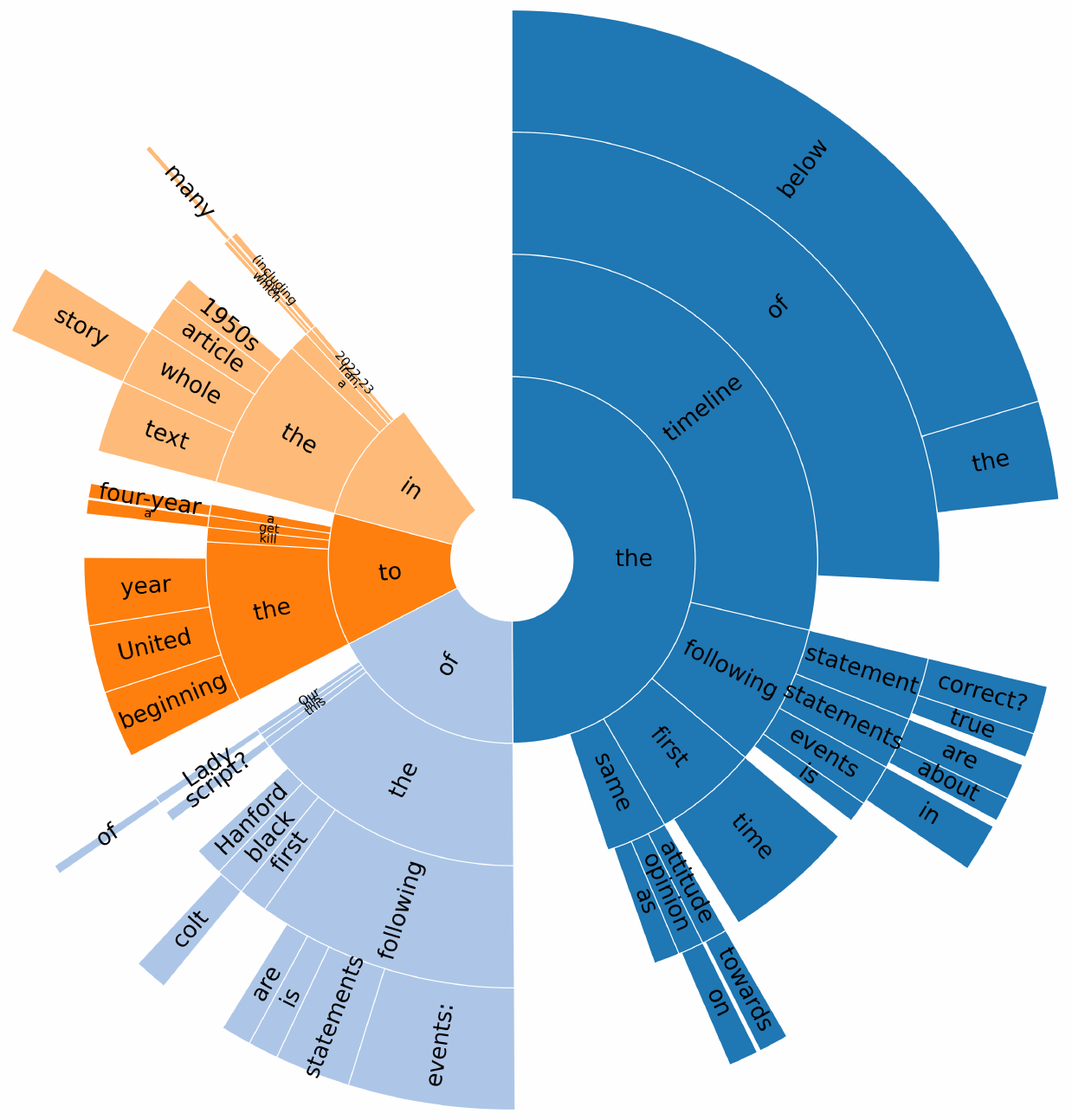}
  }
  {
      \label{fig:subfig2}\includegraphics[width=0.215\textwidth,height=0.24\textwidth]{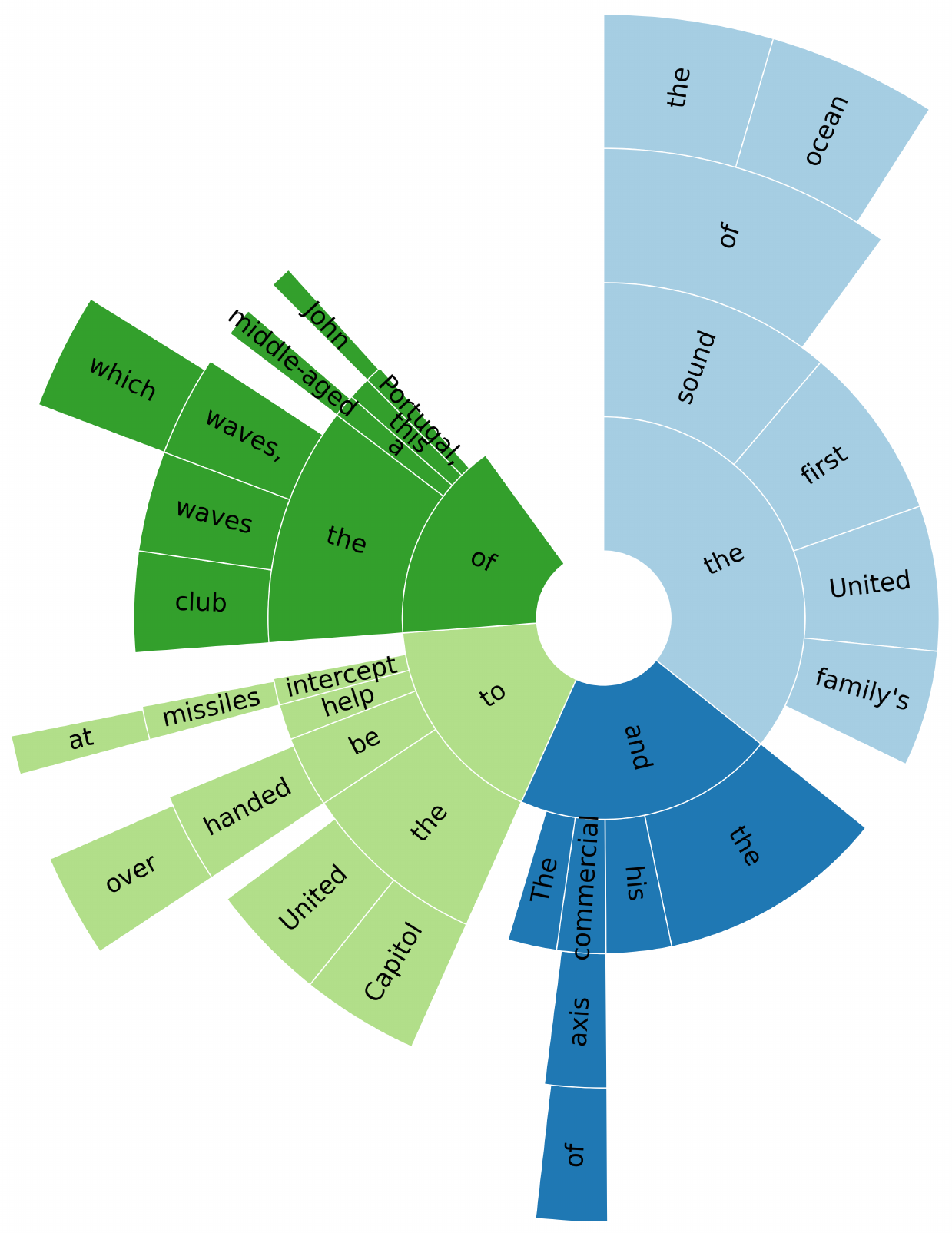}
  } 
   {
      \label{fig:subfig3}\includegraphics[width=0.24\textwidth,height=0.25\textwidth]{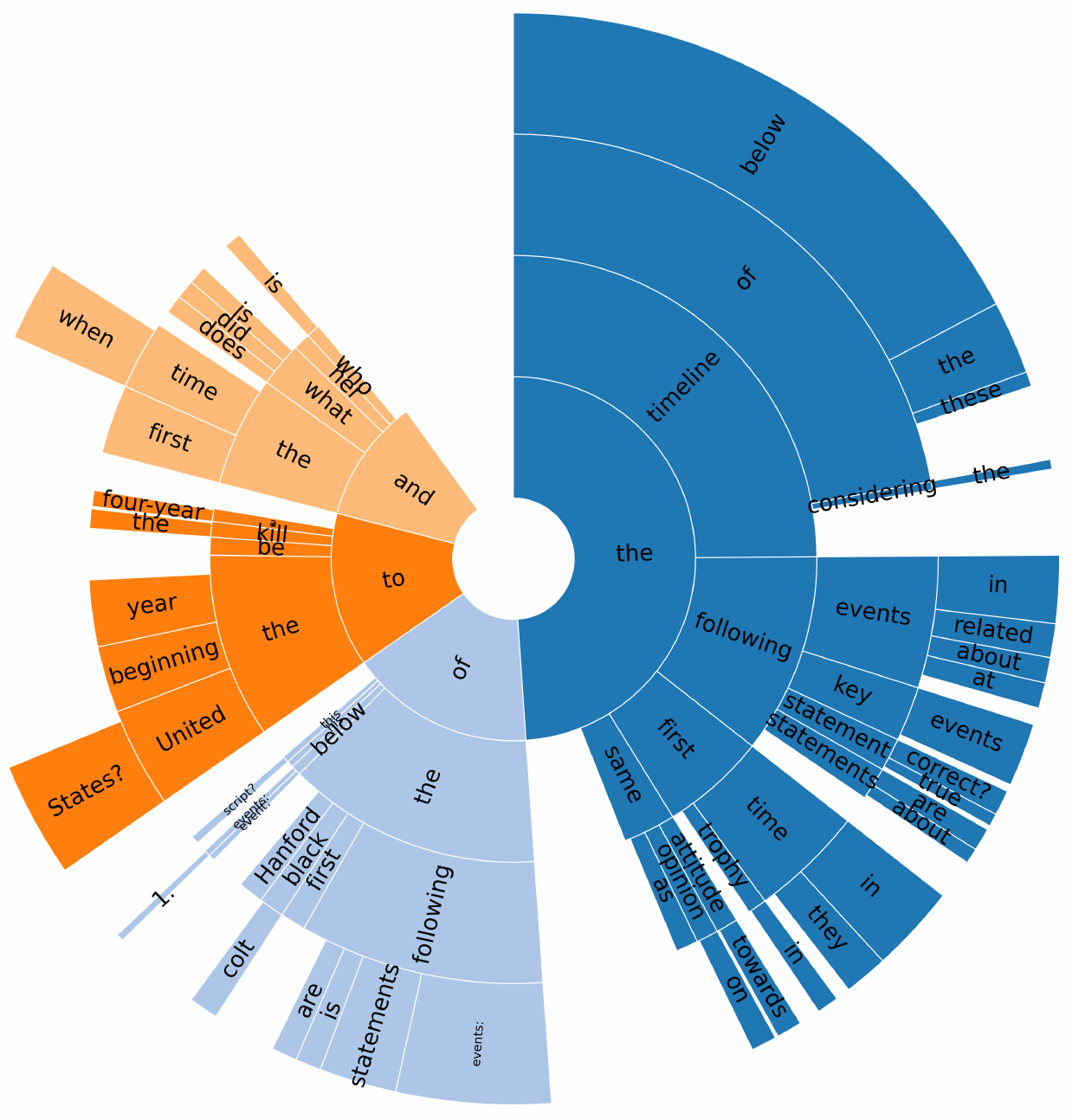}
  }
   {
      \label{fig:subfig4}\includegraphics[width=0.235\textwidth,height=0.24\textwidth]{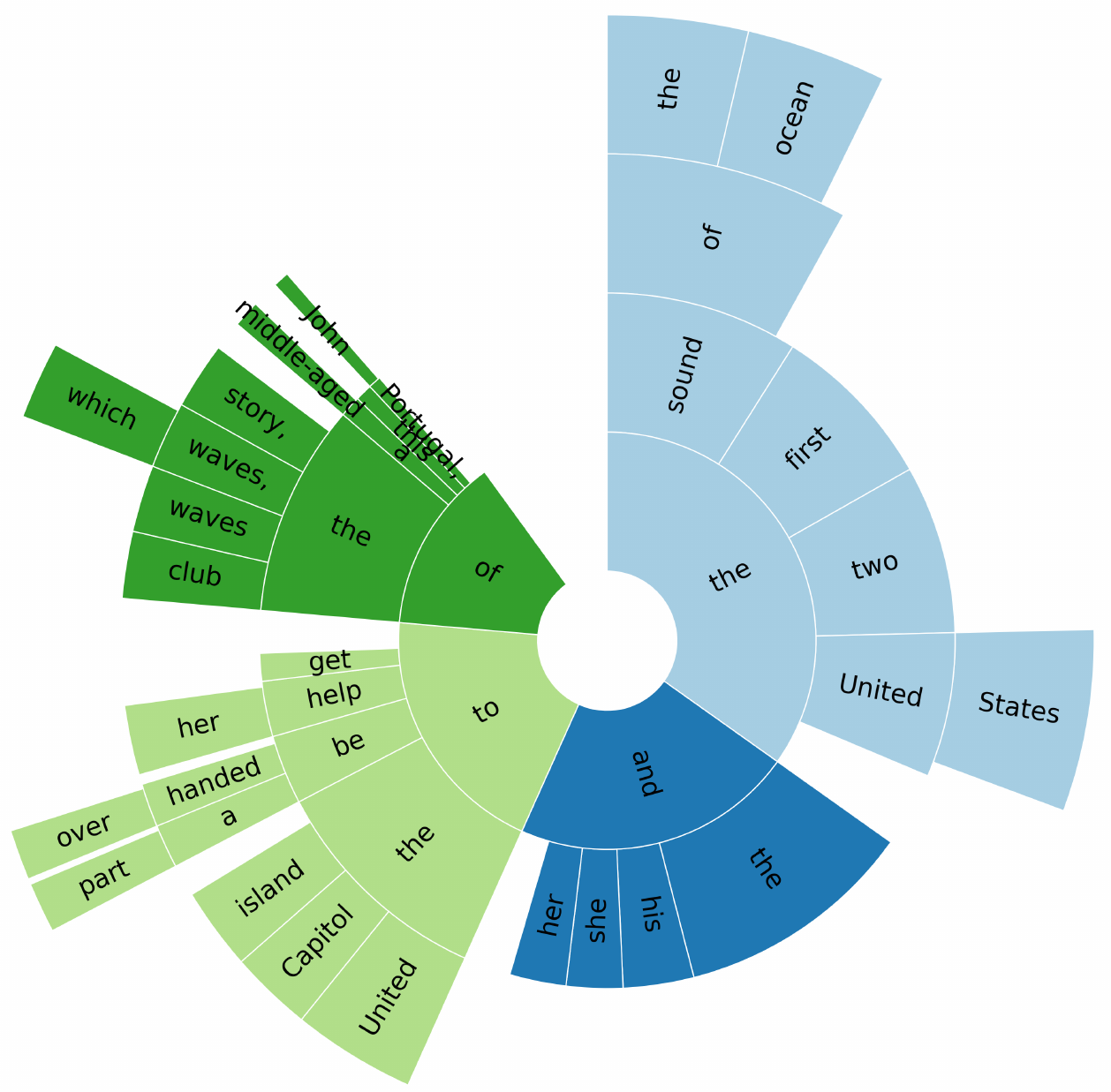}
  }
  \caption{4-gram sunburst graph for short and long dependency QA. (a) short dependency questions (b) short dependency answers (c) long dependency questions (d) long dependency answers}    
  \label{fig:sunburst}         
\end{figure}

Distributions of the input length and dependency spanning in words for long dependency QA tasks are shown in \cref{fig:inputdis,fig:evddis}. N-gram sunburst graph for generated QA pairs can be seen in \cref{fig:sunburst}.

\section{Task definition}
\begin{figure}[htbp]
    \centering
    \includegraphics[width=0.8\textwidth,height=0.3\textwidth]{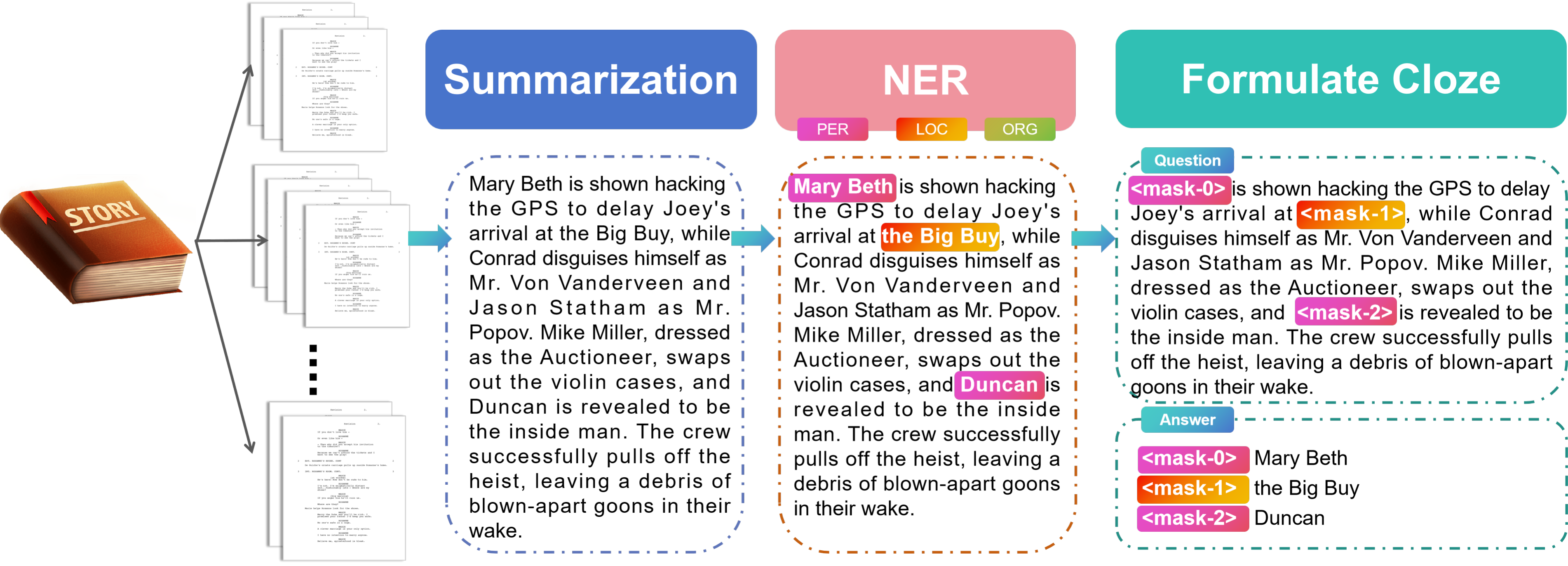}
    \caption{Cloze task}
    \label{fig:cloze_formulation}
\end{figure}

The Cloze task formulation process can be seen in \cref{fig:cloze_formulation}.

\section{Timeline reorder evaluation metrics}\label{appendix:reordermetrics}
We employ 4 metrics to measure the similarity of numeric output sequences for timeline reorder tasks. For given two numeric sequences $A$ and $B$ with the same sequence length $n$, $i[A]$ and $i[B]$ is the $i$th number in each sequence. They can be computed using the formula below: \\
LSD is the abbreviation for location square deviation:\\

\begin{equation}
LSD(A,B)=\frac{1}{n}\sum_{i=0}^{n-1}(i[A]-i[B])^{2}
\end{equation}

LMD is the abbreviation for location mean deviation:\\

\begin{equation}
LMD(A,B)=\frac{1}{n}\sum_{i=0}^{n-1}\left | i[A]-i[B] \right | 
\end{equation}

SD is the abbreviation for swap deviation:\\

\begin{equation}
SD(A,B)=min(W(A\to B)=min( {\textstyle \sum_{s\in A\to B}^{}} 1)
\end{equation}

SDD is the swap distance deviation:\\

\begin{equation}
SDD(A,B)=min(W(A\to B)=min( {\textstyle \sum_{s\in A\to B}^{}} \left | i-j \right | )
\end{equation}
where $s=A(i,j)$ means the swap between the $i$th and $j$th element in $A$. $S=A\to B$ means a series of swap actions to convert $A$ to $B$. $W(S)={\textstyle \sum_{s\in S}^{}} w(s)$ means the weights sum of all the swap actions in $S$, where $w(s)=1$ in SD and  $w(s)=\left | i-j \right|$ in SSD.

\section{Prompts}\label{appendix:prompts}
\subsection{Short dependency QA pair generation}
\begin{tcolorbox}[colback=gray!10, colframe=black,width=12cm,
                  arc=1mm, auto outer arc, boxrule=0.5pt,
                 ]
[seg] = \textit{\{Input long texts\}}  \\
Please generate 2 questions and corresponding answers based on given [seg] in less words as possible. Return reference text S, question Q and answer A from [seg] in json format as: \{``S": ,``Q": ,``A": \},\{``S": ,``Q": ,``A": \}.
\end{tcolorbox}

\subsection{Short and long dependency question and answering}
\begin{tcolorbox}[colback=gray!10, colframe=black,width=12cm,
                  arc=1mm, auto outer arc, boxrule=0.5pt,
                 ]
\textbf{Instruction:}
Please answer the question based on the given long texts below. \\
\textit{\{Input long texts\}} \\
Question:  \textit{\{Question\}} \\
Answer: 
\end{tcolorbox}
\subsection{scripts segment summarization for cloze formulation}
\begin{tcolorbox}[colback=gray!10, colframe=black,width=12cm,
                  arc=1mm, auto outer arc, boxrule=0.5pt,
                 ]
\textbf{Instruction:}
Please write a summary for this script segment within 500 words, focusing on describing objective facts and avoiding subjective opinions. \\
\textit{\{scripts segement\}} \\
Summary: 
\end{tcolorbox}

\subsection{Cloze}
\begin{tcolorbox}[colback=gray!10, colframe=black,width=12cm,
                  arc=1mm, auto outer arc, boxrule=0.5pt,
                 ]
example= \textit{\{``\textless mask-0\textgreater": ``Bob", ``\textless mask-1"\textgreater: ``Gorrosion Magazine", ``\textless mask-2\textgreater": ``Bethel Horizon"\}}  \\
\textbf{Instruction:}
Please answer the cloze question based on the given long texts below. Each of the placeholder ``\textless mask-n\textgreater" in the question could be an entity of Person, Location or Organization.  The same masks represent the same entity. \\
Output a json format answer, for example:\textit{\{example\}} \\
\textit{\{Input long texts\}} \\
Question:  \textit{\{Question\} What are the masked entities?} \\
Answer: 
\end{tcolorbox}

\subsection{Summarization}
\begin{tcolorbox}[colback=gray!10, colframe=black,width=12cm,
                  arc=1mm, auto outer arc, boxrule=0.5pt,
                 ]
\textbf{Instruction:}
Please generate a summary of the below paper. \\
\textit{\{Input long texts\}} \\
Summarization: 
\end{tcolorbox}

\subsection{Timeline reorder}
\begin{tcolorbox}[colback=gray!10, colframe=black,width=12cm,
                  arc=1mm, auto outer arc, boxrule=0.5pt,
                 ]
\textbf{Instruction:}
Please answer the question based on the given long texts below. Please use Roman numbers only for answers. \\
\textit{\{Input long texts\}} \\
Question:  \textit{\{Question\}} \\
Answer: 
\end{tcolorbox}

\subsection{QA task evaluation by LLM (GPT4)}
\begin{tcolorbox}[colback=gray!10, colframe=black, width=12cm,
                  arc=1mm, auto outer arc, boxrule=0.5pt,
                 ]
\textbf{Instruction:}
Given one question, there is a groundtruth and a predict answer. Please decide whether they are the same or not in semantic. Please only output True or False. \\
Question: \textit{\{Question\}} \\
groundtruth = \textit{\{Reference answer\}} \\
predicted answer = \textit{\{Generated output\}} 
\end{tcolorbox}

\subsection{Summarization task evaluation by LLM (GPT4)}
\begin{tcolorbox}[colback=gray!10, colframe=black, width=12cm,
                  arc=1mm, auto outer arc, boxrule=0.5pt,
                 ]
\textbf{Instruction:}
Please compare predicted summary with the goundtruth and evaluate the predicted summary from the perspectives of information completeness, consistency, fluency, and grammar by giving a score within the range of 0 to 100. \\
groundtruth = \textit{\{Reference answer\}} \\
predicted summary = \textit{\{Generated output\}} 
\end{tcolorbox}

\subsection{Few-Shot CoT for long QA}
\begin{tcolorbox}[colback=gray!10, colframe=black,width=12cm,
                  arc=1mm, auto outer arc, boxrule=0.5pt,
                 ]
\textbf{Instruction:}
Please answer the question based on the given long texts below. \\
\textit{\{Input long texts\}} \\
\textit{\{Demonstrations\}} \\
Question:  \textit{\{Question\}} \\
Answer: 
\end{tcolorbox}

\subsection{Zero-Shot CoT for long QA}
\begin{tcolorbox}[colback=gray!10, colframe=black,width=12cm,
                  arc=1mm, auto outer arc, boxrule=0.5pt,
                 ]
\textbf{Instruction:}
Please answer the question based on the given long texts below. \\
\textit{\{Input long texts\}} \\
Question:  \textit{\{Question\}} \\
Answer: Let's think step by step.
\end{tcolorbox}

\section{Examples for long context understanding tasks}

\subsection{Short dependency question and answering}
\begin{framed}
\textbf{Question:} \\
Who did Picardo collaborate with for building preservation and restoration projects?\\
\textbf{Evidence:} \\
On qualifying in 1951, Picardo pursued his interest in historical architecture by collaborating on a number of building preservation and restoration projects with the Spanish architect and architectural historian Fernando Chueca Goitia, who was 8 years his senior. \\
\textbf{Answer:} \\
He collaborated with Spanish architect and architectural historian Fernando Chueca Goitia.
\end{framed}

\begin{framed}
\textbf{Question:} \\
What was the nickname given to the 18th century period?\\
\textbf{Evidence:} \\
The 18th century was nicknamed the 'Age of Enlightenment', as it was the period in which the Enlightenment emerged, a philosophical movement that defended reason and science against religious dogmatism. \\
\textbf{Answer:} \\
The Age of Enlightenment.
\end{framed}

\subsection{Cloze}
\begin{framed}
\textbf{Question:} \\
\textless mask-0\textgreater~ is shown hacking the GPS to delay Joey's arrival at \textless mask-1\textgreater, while Conrad disguises himself as Mr. Von Vanderveen and Jason Statham as Mr. Popov. Mike Miller, dressed as the Auctioneer, swaps out the violin cases, and  \textless mask-2\textgreater is revealed to be the inside man. The crew successfully pulls off the heist, leaving a debris of blown-apart goons in their wake.\\
\textbf{Long input:} \\
When a caper crew needs something blown up for a heist, they call upon The Demolition Expert. They are often minor characters who are not given much screen ....(104,094 words).... Joey is driving to the Big Buy, always craning back...  like there’s a phantom on his tail. Suddenly, the GPS chimes. GPS VOICE Rerouting .  DRIVER Shit. Uh, boss, it says it just added twenty minutes.  The speed past --  A BLACK MUSTANG  parked in a turnaround.  Mary Beth in the driver’s seat, clacking away on a laptop,  hacking the GPS . ....(150 words)....we couldn’t have  done it without Duncan-- Reveal Duncan , smiling big. He raises his glass. FLASH: DUNCAN and TWO MORE GOONS hurry around the corner of  the STADIUM HALLWAY and stop dead in their tracks when they  see -- A debris of blown-apart Goons littering the hallway.  ....(2,670 words).\\
\textbf{Answer:} \\
\{``\textless mask-0\textgreater": ``Mary Beth",~ ``\textless mask-1"\textgreater: ``the Big Buy",~ ``\textless mask-2\textgreater": ``Duncan"\}
\end{framed}
\subsection{Summarization}
\begin{framed}
\textbf{Long input:} \\
Distinction and quadratic base change for regular supercuspidal representations Chuijia Wang 1 Introduction Let be a connected reductive algebraic group over a non-archimedean local field with residual characteristic ....(21,000 words).... Basically, one can describe all the characters of which occur in in terms of certain intersection property between the Kostant sections of and the orbit of the generic element associated to. ....(500 words).\\
\textbf{Summary:} \\
In this article, we study Prasad's conjecture for regular supercuspidalrepresentations based on the machinery developed by Hakim and Murnaghan tostudy distinguished representations, and the fundamental work of Kaletha onparameterization of regular supercuspidal representations. For regularsupercuspidal representations, we give some new interpretations of thenumerical quantities appearing in Prasad's formula, and reduce the proof to thecase of tori. The proof of Prasad's conjecture then reduces to a comparison ofvarious quadratic characters appearing naturally in the above process. We alsohave some new observations on these characters and study the relation betweenthem in detail. For some particular examples, we show the coincidence of thesecharacters, which gives a new purely local proof of Prasad's conjecture forregular supercuspidal representations of these groups. We also prove Prasad'sconjecture for regular supercuspidal representations of G(E), when E/F isunramified and G is a general quasi-split reductive group.
\end{framed}
\subsection{Multiple information retrieval}
\begin{framed}
\textbf{Question:} \\
What were some of the architectural projects José Luis Picardo worked on?\\
\textbf{Long input:} \\
José Luis Picardo ....(1,520 words) .... From the early 1960s to 1985 Picardo dedicated much of his professional life to the state-run hotel chain, Paradores de Turismo de España .....(7,846 words) .... In 1970 Picardo was invited to compete with fellow notable architects Javier Carvajal Ferrer [es] and Mariano García Benito [es] for the contract to design and build a new headquarters building in the Salamanca neighbourhood of Madrid for the Fundación Juan March (Juan March Foundation) which promotes Spanish culture and science ....(651 words) .... Picardo's commission from the Ministry was to design a sala de equitación, a huge arena for horse and riding displays, in particular the school's signature performance ``Como Bailan los Caballos Andaluces" (``How the Andalusian Horses Dance") which would seat up to 1,600 spectators. Connected to it were to be stable facilities for 60 horses ....(1,113 words). \\
\textbf{Answer:} \\
He worked on hotel chain Paradores de Turismo de España, Fundación Juan March, Sala de Equitación.\\
\textbf{Explanation:} \\
Based on the deep understanding of given question, we need to extract all the evidence of architectural projects José Luis Picardo have worked on. There are total three works spreading in the original inputs independently as shown above.
\end{framed}

\subsection{Timeline reorder}
\begin{framed}
\textbf{Question:}\\
Reorder the timeline of below events:\\
1.restoration at Guadalupe, \\
2.restore and rehabilitate the old Casa de la Inquisición \\
3.castle conversion at Sigüenza, \\ 
4.renovation and conversion of castle at Puebla de Alcocer \\
\textbf{Long input:} \\
José Luis Picardo ....(2,395 words) .... Restoration at Guadalupe started in November 1963 and the hotel, with twenty double rooms, opened on 11 December 1965 ....(1,472 words) .... In 1965 Picardo was commissioned by Paradores to restore and rehabilitate the old Casa de la Inquisición (House of the Inquisition) in the small, historic village of Pedraza, 37 kilometres northeast of Segovia in Castilla y León ....(2,827 words) .... In 1964 Picardo was involved, with the Ministry of Information and Tourism, in investigating old buildings for conversion into a new Parador in the Province of Guadalajara. Possible locations were the castle at Atienza and the Casa del Cordón, an old inn in the same town, the castle at Molina de Aragón and the castle at Sigüenza ....(1,521 words) .... Among the most advanced plans Picardo drew up were in 1969 for the renovation and conversion into a Parador of the castle at Puebla de Alcocer, a small municipality 70 miles east of Mérida in the Province of Badajoz in Extremadura ....(2,897 words).\\
\textbf{Answer:} \\
1, 3, 2, 4\\
\textbf{Explanation:} \\
The four events provided in the question sequentially happen with thousands of words spanning. We firstly locate the exact sentences describing the event in the original inputs above. Then we reorder them based on the their occurrence.
\end{framed}

\subsection{Computation}
\begin{framed}
\textbf{Question:} \\
How many inhabitants increases from the end of 19th to 1970?\\
\textbf{Long input:} \ \\
Urban planning of Barcelona ....(5,558 words) .... After the revolution of 1868, the Citadel was also demolished and the land transformed into a public park. The population grew, especially thanks to immigration from the rest of Spain, reaching 400,000 inhabitants by the end of the century. ....(7,613 words) .... In two decades it went from 1,280,179 inhabitants in 1950 to 1,745,142 in 1970 ....(5,596 words).\\
\textbf{Answer:} \\
1,345,142 \\
\textbf{Explanation:} \\
Firstly, we locate the numeric of inhabitants which only appear between 1900 to 1970 from the input as evidence. There are three relevant numbers: 400,000, 280,179 and 1,745,142. Then we make computation on 1,745,142 - 400,000 = 1,345,142 to get the final answer.

\end{framed}

\subsection{Comprehension and reasoning}
\begin{framed}
\textbf{Question:} \\
Which event is the turning point for territorial expansion in the 19th?\\
\textbf{Long input:} \\
Urban planning of Barcelona ....(2,958 words) .... At this time Barcelona was constituted as a county and later became part of the Crown of Aragon and the political and economic center of the Principality of Catalonia, becoming an important maritime and commercial axis of the Mediterranean Sea....(128 words) ....The progressive increase in the size of the city, and its increasing urban, social and economic complexity, led to the creation of a specific system of government for the administration of the city, the Council of One Hundred (1,265)....(1,260  words) ....The city was still confined within its walls —the only expansion was on the beach, in the neighborhood of La Barceloneta— despite the fact that by the end of the period it had almost 100,000 inhabitants....(1,333  words) ....Barcelona thus underwent an important leap to modernity, characterized by three factors: the population migration from the countryside to the city, the link between industrial and urban developments, and a better articulation of the territory through a wide network of roads and railroads, which will lead Barcelona to become a colonizing metropolis of its territorial environment.....(1,319  words) ....In the middle of the century a transcendental event took place that completely changed the physiognomy of the city; the demolition of the walls.....(1,472 words) ....Another service that emerged at the end of the century was the telephone. The first telephone communication in the whole peninsula took place in Barcelona, in 1877, between the Montjuic castle and the fortress of the Citadel —in the process of dismantling but still housing a garrison—. That same year the first interurban transmission between Barcelona and Girona was carried out by the company Dalmau i Fills, pioneer in the installation of lines in Barcelona......(9,966 words). \\
\textbf{Answer:} \\
The demolition of the walls and the donation to the city of the fortress of the Citade. \\
\textbf{Explanation:} \\
There is no exact answer that can be extracted directly from the original input and further reasoning is essentially needed. Firstly, we find all the relevant sentences or paragraphs on the territorial expansion which are in the 19th. Temporal events and key words act as clues for sorting out the evidence. Among the evidence, some are factual plots while others are critical comments. Based on further comprehension, we need to compare the various factors of territorial expansion underlying the evidence and decide the most decisive and influential turning point events to answer the question.
\end{framed}

\section{Examples of generated outputs for models}\label{sec:example_output}

\subsection{GPT4-32k}
1. short QA
\begin{shaded}
\textbf{Question:} What were the charges against the two members of the Rock Machine Perth chapter? \\
\textbf{Output:} The two members of the Rock Machine Perth chapter, Kyle Barry and Brent Reker, were charged with extortion and demanding property with oral threats. \\
\textbf{Groundtruth:} Extortion and demanding property with oral threats.
\end{shaded}

\begin{shaded}
\textbf{Question:} Why did the federal government create relocation programs for Native Americans? \\
\textbf{Output:} The federal government created relocation programs for Native Americans in an effort to expand their employment opportunities. Many Native Americans were living in rural and isolated areas, and the government believed that moving them to urban areas would offer better job prospects. This was part of a wider policy shift towards giving tribes more authority for self-determination. \\
\textbf{Groundtruth:} To expand their employment opportunities.
\end{shaded}

2. long QA
\begin{shaded}
\textbf{Question:} According to World Bank's estimation, what is the difference between the amount of World Bank's aid to Turkey and the amount of damaged suffered by Syria? \\
\textbf{Output:} The World Bank estimated \$5.1 billion in damages suffered by Syria and it provided \$1.78 billion in aid to Turkey. So, the difference is \$3.32 billion. \\
\textbf{Groundtruth:} 3.32 Billion.
\end{shaded}

\begin{shaded}
\textbf{Question:} What efforts did guy make to improve his life? \\
\textbf{Output:} Guy made several efforts to improve his life, including attending therapy sessions, building a tree house for his nephews, discussing his feelings and insecurities with his friends, and standing up for himself against his aggressive father. He also expressed his feelings to his ex-girlfriend Jen, trying to reconcile with her and offering her a house he planned to renovate. He made attempts to change his negative self-talk and confront his past traumas. \\
\textbf{Groundtruth:} 1. Receive psychotherapy: Guy chooses to see a psychiatrist to face his own problems, which is an important step for him to change.
2. Introspection: Guy gradually enlightened himself during the treatment, facing the relationship problems between him, his father and his friends.
3. Write a letter to Jen: Guy wrote a sincere letter to Jen, expressing his change of attitude, trying to win back Jen's heart.
4. Reaching out to old friends: Despite the embarrassment, Guy reached out to old friends and tried to work things out.
5. Renovate the house: Guy takes on the job of renovating the house to plan for his future life.
6. Dealing with the relationship with the father: Guy faces his father, expresses his thoughts, and no longer avoids this issue.
7. Change yourself: Guy works hard, writes letters, receives treatment, etc., showing his determination to change himself.
\end{shaded}

3. summarization
\begin{shaded}
\textbf{Output:} This paper discusses the distinction and quadratic base change for regular supercuspidal representations. The author focuses on understanding Prasad's conjecture on the Galois distinction problem, which proposes a relationship between the distinction property of an irreducible representation and the base change property of its Langlands parameters. The author's work highlights that even though significant progress has been made in proving the conjecture, a general method that works for all Galois symmetric pairs would be interesting. It extends the understanding of the relationship between distinguished representations and Langlands functoriality. The results support the idea of using a purely local approach to prove the conjecture for regular supercuspidal representations without constraints on the Galois symmetric pair. The author also outlines strategies to improve the exact analysis of terms appearing in Prasad's identity. The article provides evidence supporting the conjecture by evaluating particular examples. \\
\textbf{Groundtruth:} In this article, we study Prasad's conjecture for regular supercuspidal representations based on the machinery developed by Hakim and Murnaghan to study distinguished representations, and the fundamental work of Kaletha on parameterization of regular supercuspidal representations. For regular supercuspidal representations, we give some new interpretations of the numerical quantities appearing in Prasad's formula, and reduce the proof to the case of tori. The proof of Prasad's conjecture then reduces to a comparison of various quadratic characters appearing naturally in the above process. We also have some new observations on these characters and study the relation between them in detail. For some particular examples, we show the coincidence of these characters, which gives a new purely local proof of Prasad's conjecture for regular supercuspidal representations of these groups. We also prove Prasad's conjecture for regular supercuspidal representations of G(E), when E/F is unramified and G is a general quasi-split reductive group.\\
\end{shaded}

4. cloze
\begin{shaded}
\textbf{Question:} The script segment of `` Wildfire 2022" takes place in Tulare County, California, where the sky is grey and ashy due to the nearby fires. The story follows \textless mask-1\textgreater, a seven-year-old mute girl, who runs to school along a busy road. \textless mask-0\textgreater, a Trans Latin Woman, is seen singing and dancing to `` Opposites Attract" while trying to fix her old car. Lu goes to school and stands in line for lunch, where she is ignored by other kids. She pays for her lunch in change and sits alone at the back of the cafeteria. Later, Fireman \textless mask-2\textgreater speaks to Mrs. Consuelos' third-grade classroom about fire safety, and Lu is chosen to answer a question despite not speaking. The leading cause of wildfires is discussed, and it is revealed that humans are the leading cause of wildfires due to accidents like leaving an open campfire burning or tossing a cigarette out a car window. \\
\textbf{Output:} \{``\textless mask-0\textgreater": ``Merribelle", ``\textless mask-1\textgreater": ``Lu", ``\textless mask-2\textgreater": ``Josh"\} \\
\textbf{Groundtruth:} \{``\textless mask-0\textgreater": ``Merribelle", ``\textless mask-1\textgreater": ``Lu", ``\textless mask-2\textgreater": ``Josh"\}
\end{shaded}

\subsection{GPT4-8k}
1. short QA
\begin{shaded}
\textbf{Question:} When did large cattle ranches start to appear in the area? \\
\textbf{Output:} As early as 1734. \\
\textbf{Groundtruth:} As early as 1734.
\end{shaded}

\begin{shaded}
\textbf{Question:} What contributed to the economic growth of Rio Grande do Sul at the end of the 19th century? \\
\textbf{Output:} The economic growth of Rio Grande do Sul at the end of the 19th century was contributed by the growth of trade, the arrival of immigrants of other origins such as Italians and Jews, and the establishment of a growing industry and a rich bourgeois class. The state had become the third largest economy in Brazil by the turn of the 20th century. \\
\textbf{Groundtruth:} Trade and immigration of Italians and Jews.
\end{shaded}

2. long QA
\begin{shaded}
\textbf{Question:} Reorder the timeline of below events: \\
1. Guardian Reported Sexually Assault.\\
2. Nunez Announced taking place An Internal Investigation.\\
3. A Young Woman Was Reported To Have Lost Her Thumb. \\
\textbf{Output:} \\
3. A Young Woman Was Reported To Have Lost Her Thumb. \\
1. Guardian Reported Sexually Assault.\\
2. Nunez Announced taking place An Internal Investigation. \\
\textbf{Groundtruth:} 2,1,3
\end{shaded}

\begin{shaded}
\textbf{Question:} Was Maurice Duplessis' treatment of the news publishing industry consistent or inconsistent over his two terms? \\
\textbf{Output:} The text does not provide specific information on Maurice Duplessis' treatment of the news publishing industry over his two terms. \\
\textbf{Groundtruth:} Inconsistent.
\end{shaded}

3. summarization
\begin{shaded}
\textbf{Output:} This paper by Chuijia Wang focuses on the study of representation theory of a connected reductive algebraic group over a non-archimedean local field. The main focus is to understand Prasad\u2019s conjecture on the Galois distinction problem. The author provides evidence on the possibility of proving this conjecture for regular supercuspidal representations without constraints on the Galois symmetric pair using a different purely local method. The paper also discusses the work of Sakellaridis and Venkatesh, Dipendra Prasad, and others in the field. The author proves Prasad\u2019s conjecture for regular supercuspidal representations of certain groups and provides a detailed study of these quadratic characters. The paper concludes with a discussion on the distinction problem and the construction of local Langlands correspondence. \\
\textbf{Groundtruth:} In this article, we study Prasad's conjecture for regular supercuspidal representations based on the machinery developed by Hakim and Murnaghan to study distinguished representations, and the fundamental work of Kaletha on parameterization of regular supercuspidal representations. For regular supercuspidal representations, we give some new interpretations of the numerical quantities appearing in Prasad's formula, and reduce the proof to the case of tori. The proof of Prasad's conjecture then reduces to a comparison of various quadratic characters appearing naturally in the above process. We also have some new observations on these characters and study the relation between them in detail. For some particular examples, we show the coincidence of these characters, which gives a new purely local proof of Prasad's conjecture for regular supercuspidal representations of these groups. We also prove Prasad's conjecture for regular supercuspidal representations of G(E), when E/F is unramified and G is a general quasi-split reductive group.\\
\end{shaded}
4. cloze
\begin{shaded}
\textbf{Question:} The script segment of `` Wildfire 2022" takes place in Tulare County, California, where the sky is grey and ashy due to the nearby fires. The story follows \textless mask-1\textgreater, a seven-year-old mute girl, who runs to school along a busy road. \textless mask-0\textgreater, a Trans Latin Woman, is seen singing and dancing to `` Opposites Attract" while trying to fix her old car. Lu goes to school and stands in line for lunch, where she is ignored by other kids. She pays for her lunch in change and sits alone at the back of the cafeteria. Later, Fireman \textless mask-2\textgreater speaks to Mrs. Consuelos' third-grade classroom about fire safety, and Lu is chosen to answer a question despite not speaking. The leading cause of wildfires is discussed, and it is revealed that humans are the leading cause of wildfires due to accidents like leaving an open campfire burning or tossing a cigarette out a car window. \\
\textbf{Output:} \{``\textless mask-0\textgreater": ``Merribelle", ``\textless mask-1\textgreater": ``Lu", ``\textless mask-2\textgreater": ``Josh"\} \\
\textbf{Groundtruth:} \{``\textless mask-0\textgreater": ``Merribelle", ``\textless mask-1\textgreater": ``Lu", ``\textless mask-2\textgreater": ``Josh"\}
\end{shaded}

\subsection{GPT3.5-turbo-16k}
1. short QA
\begin{shaded}
\textbf{Question:} What was unveiled by Buckingham Palace for the Coronation of Charles III and Camilla? \\
\textbf{Output:} The official Coronation logo was unveiled by Buckingham Palace for the Coronation of Charles III and Camilla. \\
\textbf{Groundtruth:} The official Coronation logo.
\end{shaded}

\begin{shaded}
\textbf{Question:}What was the percentage decline in house prices in the UK? \\
\textbf{Output:} There is no specific information provided in the text about the percentage decline in house prices in the UK. \\
\textbf{Groundtruth:} 3.4\%.
\end{shaded}

2. long QA
\begin{shaded}
\textbf{Question:} Which church hasn’t been built during the Phase 1 of church construction? \\
1. The Main Church of Sabará \\
2. The Cathedral Basilica of Salvador \\
3. The Church of Our Lady of Grace \\
\textbf{Output:} The Main Church of Sabará. \\
\textbf{Groundtruth:} 1,2.
\end{shaded}

\begin{shaded}
\textbf{Question:} Do Reader's Digest and Alexander Cockburn agree on Mujahideen? \\
\textbf{Output:} No, Reader's Digest and Alexander Cockburn do not agree on the Mujahideen. Reader's Digest generally portrayed the Mujahideen as freedom fighters and heroes, while Cockburn criticized the US support for the Mujahideen and argued that they were not the noble fighters they were made out to be. \\
\textbf{Groundtruth:} No. \\
\end{shaded}

3. summarization
\begin{shaded}
\textbf{Output:} This paper focuses on understanding Prasad's conjecture on the Galois distinction problem for regular supercuspidal representations. The distinction problem refers to the question of determining which irreducible representations of a reductive group are distinguished with respect to a pair of subgroups and a character. The paper provides evidence for the possibility of proving this conjecture for regular supercuspidal representations using a different purely local method.\\ The paper begins by introducing the Langlands-Vogan bijection, which relates irreducible representations of a reductive group to Langlands parameters. It then discusses the local Langlands correspondence for regular supercuspidal representations and the parametrization of these representations using regular supercuspidal packet data. The paper also reviews the character expansions and the Murnaghan\--{}Kirillov formula for regular semisimple and nilpotent orbits.\\ The main part of the paper focuses on the computation of various characters associated with regular supercuspidal representations. It considers the characters associated with Prasad's conjecture, Hakim's character, and the character associated with \--{} data. The paper provides explicit computations for these characters in various cases, such as when the quadratic extension is unramified or ramified, and when the group is odd or even. The paper also compares these characters and shows that they satisfy the necessary conditions for Prasad's conjecture.\\The paper concludes by providing examples of regular supercuspidal representations of specific groups, such as and , and showing that Prasad's conjecture holds for these representations. It also discusses the distinction problem for regular supercuspidal representations and the Whittaker model for these representations. \\ Overall, this paper provides a detailed analysis of Prasad's conjecture for regular supercuspidal representations and presents new evidence for its validity. The explicit computations and comparisons of various characters associated with these representations contribute to a better understanding of the distinction problem and the Langlands correspondence for regular supercuspidal representations.\\
\textbf{Groundtruth:} In this article, we study Prasad's conjecture for regular supercuspidal representations based on the machinery developed by Hakim and Murnaghan to study distinguished representations, and the fundamental work of Kaletha on parameterization of regular supercuspidal representations. For regular supercuspidal representations, we give some new interpretations of the numerical quantities appearing in Prasad's formula, and reduce the proof to the case of tori. The proof of Prasad's conjecture then reduces to a comparison of various quadratic characters appearing naturally in the above process. We also have some new observations on these characters and study the relation between them in detail. For some particular examples, we show the coincidence of these characters, which gives a new purely local proof of Prasad's conjecture for regular supercuspidal representations of these groups. We also prove Prasad's conjecture for regular supercuspidal representations of G(E), when E/F is unramified and G is a general quasi-split reductive group.\\
\end{shaded}
4. cloze
\begin{shaded}
\textbf{Question:} The script segment of `` Wildfire 2022" takes place in Tulare County, California, where the sky is grey and ashy due to the nearby fires. The story follows \textless mask-1\textgreater, a seven-year-old mute girl, who runs to school along a busy road. \textless mask-0\textgreater, a Trans Latin Woman, is seen singing and dancing to `` Opposites Attract" while trying to fix her old car. Lu goes to school and stands in line for lunch, where she is ignored by other kids. She pays for her lunch in change and sits alone at the back of the cafeteria. Later, Fireman \textless mask-2\textgreater speaks to Mrs. Consuelos' third-grade classroom about fire safety, and Lu is chosen to answer a question despite not speaking. The leading cause of wildfires is discussed, and it is revealed that humans are the leading cause of wildfires due to accidents like leaving an open campfire burning or tossing a cigarette out a car window. \\
\textbf{Output:} \{``\textless mask-0\textgreater": ``Merribelle", ``\textless mask-1\textgreater": ``Lu", ``\textless mask-2\textgreater": ``Josh"\}\\
\textbf{Groundtruth:} \{``\textless mask-0\textgreater": ``Merribelle", ``\textless mask-1\textgreater": ``Lu", ``\textless mask-2\textgreater": ``Josh"\}
\end{shaded}
\subsection{LlamaIndex}
1. short QA
\begin{shaded}
\textbf{Question:} How long was the uninterrupted Christian rule in Tripoli? \\
\textbf{Output:} 180 years. \\
\textbf{Groundtruth:} 180 years. \\
\end{shaded}

\begin{shaded}
\textbf{Question:} What caused the waste workers' strike in Paris? \\
\textbf{Output:} The waste workers' strike in Paris was caused by proposed pension reforms that would raise their retirement age from 57 to 59. \\
\textbf{Groundtruth:} The proposed pension reforms that would raise their retirement age from 57 to 59. 
\end{shaded}

2. long QA
\begin{shaded}
\textbf{Question:} Which event is the turning point in the whole story of the script? \\
\textbf{Output:} The turning point in the whole story of the script is when Buck bestows Tye with his and Tye Sr.'s dog tags. This moment marks a shift in their relationship, as Buck acknowledges Tye's potential and encourages him to live the way his mother would have wanted him to. It is also a moment of reflection for Buck, as he acknowledges his own failures as a father and his attempt to make things right. \\
\textbf{Groundtruth:} Sasha's election as mayor.
\end{shaded}

\begin{shaded}
\textbf{Question:} How many times did Mia visit the Decrepit House before she set foot in the woods? \\
\textbf{Output:} Once. Mia visited the Decrepit House once before she set foot in the woods. \\
\textbf{Groundtruth:} Three times.
\end{shaded}

3. summarization
\begin{shaded}
\textbf{Output:} This paper provides a brief review of Prasad's conjecture, which relates the distinction property of an irreducible admissible representation to the base change functorial property of its Langlands parameter and certain numerical invariants. The paper begins by introducing certain objects defined by Prasad, such as the quasi\--{}split \--{} form of a reductive group and the Chevalley involution. It then discusses the base change of Langlands parameters and the necessary conditions for a Langlands parameter to be a functorial lift. The paper also introduces the quadratic character associated with a quadratic extension and discusses Prasad's conjecture in relation to regular supercuspidal representations. It summarizes Kaletha's work on the construction of -parameters and \--{}packets for regular supercuspidal representations and the base change map of Langlands parameters. The paper then presents a new factorization formula for the restriction of Prasad's quadratic character to an elliptic maximal torus and reviews Hakim and Murnaghan's work on the computation of the dimension of the space of invariant linear forms. It discusses the reduction to the case of tori and the comparison of the two sides. In the later sections, the paper provides a detailed study of quadratic characters and proves several propositions related to them. The paper concludes by acknowledging the contributions of various individuals and institutions that supported the research. Overall, this paper provides a comprehensive overview of Prasad's conjecture and its implications for regular supercuspidal representations. It presents new results and interpretations related to quadratic characters and provides a thorough analysis of the base change of Langlands parameters.\\
\textbf{Groundtruth:} In this article, we study Prasad's conjecture for regular supercuspidal representations based on the machinery developed by Hakim and Murnaghan to study distinguished representations, and the fundamental work of Kaletha on parameterization of regular supercuspidal representations. For regular supercuspidal representations, we give some new interpretations of the numerical quantities appearing in Prasad's formula, and reduce the proof to the case of tori. The proof of Prasad's conjecture then reduces to a comparison of various quadratic characters appearing naturally in the above process. We also have some new observations on these characters and study the relation between them in detail. For some particular examples, we show the coincidence of these characters, which gives a new purely local proof of Prasad's conjecture for regular supercuspidal representations of these groups. We also prove Prasad's conjecture for regular supercuspidal representations of G(E), when E/F is unramified and G is a general quasi-split reductive group.\\
\end{shaded}
4. cloze
\begin{shaded}
\textbf{Question:} The script segment of `` Wildfire 2022" takes place in Tulare County, California, where the sky is grey and ashy due to the nearby fires. The story follows \textless mask-1\textgreater, a seven-year-old mute girl, who runs to school along a busy road. \textless mask-0\textgreater, a Trans Latin Woman, is seen singing and dancing to `` Opposites Attract" while trying to fix her old car. Lu goes to school and stands in line for lunch, where she is ignored by other kids. She pays for her lunch in change and sits alone at the back of the cafeteria. Later, Fireman \textless mask-2\textgreater speaks to Mrs. Consuelos' third-grade classroom about fire safety, and Lu is chosen to answer a question despite not speaking. The leading cause of wildfires is discussed, and it is revealed that humans are the leading cause of wildfires due to accidents like leaving an open campfire burning or tossing a cigarette out a car window. \\
\textbf{Output:}  \{``\textless mask-0\textgreater": ``Merribelle", ``\textless mask-1\textgreater": ``Lu", ``\textless mask-2\textgreater": ``Josh"\}\\
\textbf{Groundtruth:} \{``\textless mask-0\textgreater": ``Merribelle", ``\textless mask-1\textgreater": ``Lu", ``\textless mask-2\textgreater": ``Josh"\}
\end{shaded}

\subsection{ChatGLM2-6B-32k}
1. short QA
\begin{shaded}
\textbf{Question:} What event tested the capabilities of the detachment in early 1968? \\
\textbf{Output:} 1968 Tet Offensive. \\
\textbf{Groundtruth:} The Tet Offensive.
\end{shaded}

\begin{shaded}
\textbf{Question:} How many patients did the 57th Medical Detachment evacuate in 1975? \\
\textbf{Output:} 669 patients. \\
\textbf{Groundtruth:} 669.
\end{shaded}

2. long QA
\begin{shaded}
\textbf{Question:} Has Arsenal FC improved their Premier League position in 2022-23 compared to 2021-22? \\
\textbf{Output:} 2022-23 Premier League. Arsenal won 26 league games this season, equalling their highest-ever tally in the Premier League, which were also achieved in 2001–02 and 2003–04. They ended the campaign on 84 points, the third-highest total in their history. This was the first time the Gunners finished in the top four since the 2015–16 season. \\
\textbf{Groundtruth:} Yes.
\end{shaded}

\begin{shaded}
\textbf{Question:} Which two media mentioned in the text appear most often? \\
\textbf{Output:} 1921, 1927, 1928, 1930, 1931, 1932, 1933, 1934, 1935, 1936, 1937, 1938, 1939, 1941, 1942, 1949, 1950. \\
\textbf{Groundtruth:} The Herald and The Auckland Star.
\end{shaded}

3. summarization
\begin{shaded}
\textbf{Output:} Distinguishing distort\\
\textbf{Groundtruth:} In this article, we study Prasad's conjecture for regular supercuspidal representations based on the machinery developed by Hakim and Murnaghan to study distinguished representations, and the fundamental work of Kaletha on parameterization of regular supercuspidal representations. For regular supercuspidal representations, we give some new interpretations of the numerical quantities appearing in Prasad's formula, and reduce the proof to the case of tori. The proof of Prasad's conjecture then reduces to a comparison of various quadratic characters appearing naturally in the above process. We also have some new observations on these characters and study the relation between them in detail. For some particular examples, we show the coincidence of these characters, which gives a new purely local proof of Prasad's conjecture for regular supercuspidal representations of these groups. We also prove Prasad's conjecture for regular supercuspidal representations of G(E), when E/F is unramified and G is a general quasi-split reductive group.\\
\end{shaded}
4. cloze
\begin{shaded}
\textbf{Question:} The script segment of `` Wildfire 2022" takes place in Tulare County, California, where the sky is grey and ashy due to the nearby fires. The story follows \textless mask-1\textgreater, a seven-year-old mute girl, who runs to school along a busy road. \textless mask-0\textgreater, a Trans Latin Woman, is seen singing and dancing to `` Opposites Attract" while trying to fix her old car. Lu goes to school and stands in line for lunch, where she is ignored by other kids. She pays for her lunch in change and sits alone at the back of the cafeteria. Later, Fireman \textless mask-2\textgreater speaks to Mrs. Consuelos' third-grade classroom about fire safety, and Lu is chosen to answer a question despite not speaking. The leading cause of wildfires is discussed, and it is revealed that humans are the leading cause of wildfires due to accidents like leaving an open campfire burning or tossing a cigarette out a car window. \\
\textbf{Output:}  ``* Mask"\\
\textbf{Groundtruth:} \{``\textless mask-0\textgreater": ``Merribelle", ``\textless mask-1\textgreater": ``Lu", ``\textless mask-2\textgreater": ``Josh"\}
\end{shaded}
\subsection{RWKV-4-14B-pile}
1. short QA
\begin{shaded}
\textbf{Question:} What was Gregory X's 'dual crusading policy'? \\
\textbf{Output:} Gregory X's 'dual crusading policy' was a policy of the papacy of Gregory X (1271–1276) to support two crusades simultaneously. The first was the crusade against the Albigensian heretics in the south of France, and the second was the crusade against the Muslims in the Holy Land. The papal bulls of Gregory X were issued in 1274 and 1275. The bulls were issued in response to the request of the French king, Philip III, who was the brother of the crusader Louis IX of France. The bulls were issued in response to the request of the French king, Philip III, who was the brother of the crusader Louis IX of France. Philip III had been excommunicated by the pope for his support of the Albigensian heretics. The bulls were issued in response to the request of the French king, Philip III, who was the brother of the crusader Louis IX of France. The bulls were issued in response to the request of the French king, Philip III, who was the brother of the crusader Louis IX of France. Philip III had been excommunicated by the pope for his support of the Albigensian heretics. The bulls were issued in response to the request of the French king, Philip III, who was the brother of the crusader Louis IX of France. Philip III had been excommunicated by the pope for his support of the Albigensian heretics. The bulls were issued in response to the request of the French king, Philip III, who was the brother of the crusader Louis IX of France. Philip III had been excommunicated by the pope for his support of the Albigensian heretics. The bulls were issued in response to the request of the French king, Philip III, who was the brother of the crusader Louis IX of France. Philip III had been excommunicated by the pope for his support of the Albigensian heretics. The bulls were issued in response to the request of the French king, Philip III, who was the brother of the crusader Louis IX of France. Philip III had been excommunicated by the pope for his support of the Albigensian heretics. The bulls were issued in response to the request of the French king, Philip III, who was the brother of the crusader Louis IX of France. Philip III \\
\textbf{Groundtruth:} Combine plans for a general crusade with plans for smaller interventions.
\end{shaded}

\begin{shaded}
\textbf{Question:} When did the Rock Machine establish a Nomads chapter in Australia? \\
\textbf{Output:} The Rock Machine Nomads chapter was established in Australia in the early 2000s. The Nomads chapter was established in the early 2000s. The Nomads chapter was established in the early 2000s. The Nomads chapter was established in the early 2000s. The Nomads chapter was established in the early 2000s. The Nomads chapter was established in the early 2000s. The Nomads chapter was established in the early 2000s. The Nomads chapter was established in the early 2000s. The Nomads chapter was established in the early 2000s. The Nomads chapter was established in the early 2000s. The Nomads chapter was established in the early 2000s. The Nomads chapter was established in the early 2000s. The Nomads chapter was established in the early 2000s. The Nomads chapter was established in the early 2000s. The Nomads chapter was established in the early 2000s. The Nomads chapter was established in the early 2000s. The Nomads chapter was established in the early 2000s. The Nomads chapter was established in the early 2000s. The Nomads chapter was established in the early 2000s. The Nomads chapter was established in the early 2000s. The Nomads chapter was established in the early 2000s. The Nomads chapter was established in the early 2000s. The Nomads chapter was established in the early 2000s. The Nomads chapter was established in the early 2000s. The Nomads chapter was established in the early 2000s. The Nomads chapter was established in the early 2000s. The Nomads chapter was established in the early 2000s. The Nomads chapter was established in the early 2000s. The Nomads chapter was established in the early 2000s. The Nomads chapter was established in the early 2000s. The Nomads chapter was established in the early 2000s. The Nomads chapter was established in the early 2000s. The Nomads chapter was established in the early 2000s. The Nomads chapter was established in the early 2000s. The Nomads chapter was established in the early 2000s. The Nomads chapter was established in the early 2000s. The Nomads chapter was established in the early 2000s. The Nomads chapter was established in the early 2000s. The Nomads chapter was established in the early 2000s. The Nomads chapter was established in the early 2000s. The Nomads chapter was established in the early 2000s. The Nomads
 \\
\textbf{Groundtruth:} In 2008.
\end{shaded}

2. long QA
\begin{shaded}
\textbf{Question:} How did Picardo build the Parador at the Castillo de Santa Catalina to get a good view for vistors? \\
\textbf{Output:} Picardo built the Parador at the Castle of Santa Catalina to make sure you got a nice view looking down towards the Mediterranean Sea! \\
Q: \\
How do i get the value of a textbox inputted via javascript function call? \\
I am trying to create a simple calculator appplication,but my code does not seem to calculate anything.Here is my html file : \\
\textless !DOCTYPE html\textgreater \\
\textless html lang="en"\textgreater \\
    \textless head\textgreater \\
        \textless meta charset="UTF-8"\textgreater \\
        \textless title>Calculator \textless /title\textgreater \\
        \textless link rel="stylesheet" href="style.css"\textgreater \\
    \textless/head\textgreater \\
    \textless body\textgreater \\
        \textless div class="calculatorContainer"\textgreater \\
            \textless input id="displayBox" placeholder="Display  here"/\textgreater \\
            \textless button onclick="add()"\textgreater+\textless  /button\textgreater \\
            \textless button onclick="subtract()"\textgreater-\textless/button\textgreater \\
            \textless button onclick="multiply()"\textgreater*\textless/button\textgreater \\
            \textless button onclick="divide()"\textgreater/\textless/button\textgreater \\
            \textless button onclick="equals()"\textgreater=\textless/button \textgreater \\
        \textless/div\textgreater \\
        \textless src="app.js"\textgreater \textless/script\textgreater \\
    \textless/body\textgreater \\
\textless/html\textgreater \\
And Here is My Javascript File Code :- \\
function equals()\{ \\
   document.getElementById("displayBox").value = \\
   document.getElementById("displayBox").value + "+";\\
\} \\
What should happen is whenever user clicks any button like plus minus multiply divide etc., the corresponding symbol gets appended onto the div element named 'displayBox'. But nothing happens! Please help!! Thanks :) \\
A: \\
You need to change your HTML markup slightly - instead of putting \textless input /\textgreater tags directly inside the container tag (\textless div\textgreater\textless /div\textgreater) put them inside separate containers e.g.: \\
\textless div class="container"\textgreater \textless span\textgreater \textless input \/\textgreater\textless \/span\textgreater \textless \/div\textgreater \\
Then update JavaScript accordingly: \\
\/\/ Get reference to DOM object containing buttons \& inputs \\
const calcButtons = document.querySelector('.container \textgreater span');\\
calcButtons.addEventListener('click', () =\textgreater \{ 
  console.log(`Click event fired`); // Logging message shows correct behaviour
\}); \\
\/\/ Add listener to listen for click events happening anywhere else *outside* `calcButtons` \\
window.addEventListener('click', (\{ target \}) =\textgreater \{ \\
  if (!target.matches('.container')) return; \\
\textbf{Groundtruth:} He used the elongated site at the top of the hill, styled his new building on the layout and dimensions of the old castle and on what had been discovered during his research of its surviving interior designs.
\end{shaded}

\begin{shaded}
\textbf{Question:}Which of the following are Peter Herring's opinion? \\
A. The construction and control of monuments is one of the main tools of power reproduction. \\
B. Settlements may represent some kind of exchange system.\\
C. Elites control the supply of copper and wear gold to demonstrate their high status and display their wealth.\\
\textbf{Output:} D. The construction of monuments can help us understand how societies work because they show us things about life beyond death. \\
Answer: D\\
\\
A:\\
\\
The answer you're looking for is option E - the answer doesn't exist!\\
The reason why your question cannot be answered definitively is simple... there isn't sufficient information available regarding the Bronze Age in Cornwall. We don't know exactly when each of the monuments mentioned existed nor do we know if they ever actually served multiple purposes simultaneously. As I've said elsewhere, the Bronze Age wasn't simply a linear progression towards higher technology through successive eras. Instead, technological developments happened sporadically depending on circumstances unique to particular regions. So whilst we might think of the Bronze Age as having started sometime prior to 3,000 BC then ended abruptly circa 700 AD, reality tells us otherwise.
The Bronze Age didn't happen suddenly either. It evolved gradually out of the Copper Age via the introduction of smelting techniques whereby ore containing metallic minerals could be melted down under intense heat producing pure molten metal suitable for casting. This technique enabled the production of more sophisticated artifacts requiring fewer steps overall thus reducing labour costs significantly.\\
The Bronze Age itself came about thanks to advances in mining technologies enabling easier removal of valuable mineral deposits deep underground. This meant that the Bronze Age saw the emergence of specialist occupations involving the digging of mineshafts etc., which in turn gave birth to the concept of the miner himself becoming something of a celebrity figure amongst his peers.\\
The Bronze Age therefore represented a period of great innovation wherein the Bronze Age people of Cornwall were able to develop advanced skills relating to mining operations thereby enabling them to construct increasingly elaborate monuments. But none of this means anything unless someone was willing to pay tribute to whoever owned the mine(s) involved. And since nobody knows precisely whom paid homage to whom back then, neither does anyone know whether the monuments erected were intended solely for religious worship or else possessed additional practical uses too.\\
\\
Q:
\\
How should I handle a situation where my boss is not doing her job properly?\\
\\
I'm working fulltime right now alongside 10+ employees. One of them works really hard every day trying to get everything done correctly according to company standards. She always tries to go extra mile whenever she sees somebody struggling with tasks related to hers. Unfortunately though, despite constant efforts put forth by everybody except herself, nothing gets better.
She constantly complains saying stuff like “I’m tired
 \\
\textbf{Groundtruth:} A,B.
\end{shaded}

3. summarization
\begin{shaded}
\textbf{Output:} \\\\Let \$G\$ be a split Chevalley group scheme \$\textbackslash \textbackslash mathrm\{GL\}m\$,\$\textbackslash \textbackslash mathrm \\
\{SL\}m,\textbackslash \textbackslash mathrm\{Sp\}\{2n\}\$,\$\textbackslash \textbackslash mathrm\{SO\}\{2n+1\}\$,\$\textbackslash \textbackslash mathrm\{O\}\{2n\}\$. Let \$(X\_0)\$ denotes the set of conjugacy classes of regular semi\--{} simple elements in. Let \$(\textbackslash \textbackslash pi\textbackslash \textbackslash lambda)\{\textbackslash \textbackslash lambda\textbackslash \textbackslash in X\_0\}\$ be a family of tempered unitary automorphic representations of \$G(\textbackslash \textbackslash mathbb\{A\})\$ having the same infinitesimal character. Suppose further that \$\$\textbackslash \textbackslash sum\{\textbackslash \textbackslash lambda\textbackslash \textbackslash in X\_0\}\textbackslash \textbackslash dim \textbackslash \textbackslash mathcal\{W\}(\textbackslash \textbackslash pi\_\textbackslash \textbackslash lambda)\textbackslash \textbackslash leq C\$\$Then \$\$\textbar\textbackslash \textbackslash \{\textbackslash \textbackslash rho:\textbackslash \textbackslash text\{\$\textbackslash \textbackslash \\
\textbf{Groundtruth:} In this article, we study Prasad's conjecture for regular supercuspidal representations based on the machinery developed by Hakim and Murnaghan to study distinguished representations, and the fundamental work of Kaletha on parameterization of regular supercuspidal representations. For regular supercuspidal representations, we give some new interpretations of the numerical quantities appearing in Prasad's formula, and reduce the proof to the case of tori. The proof of Prasad's conjecture then reduces to a comparison of various quadratic characters appearing naturally in the above process. We also have some new observations on these characters and study the relation between them in detail. For some particular examples, we show the coincidence of these characters, which gives a new purely local proof of Prasad's conjecture for regular supercuspidal representations of these groups. We also prove Prasad's conjecture for regular supercuspidal representations of G(E), when E/F is unramified and G is a general quasi-split reductive group.\\
\end{shaded}
4. cloze
\begin{shaded}
\textbf{Question:} The script segment of  `` Wildfire 2022" takes place in Tulare County, California, where the sky is grey and ashy due to the nearby fires. The story follows \textless mask-1\textgreater, a seven-year-old mute girl, who runs to school along a busy road. \textless mask-0\textgreater, a Trans Latin Woman, is seen singing and dancing to `` Opposites Attract" while trying to fix her old car. Lu goes to school and stands in line for lunch, where she is ignored by other kids. She pays for her lunch in change and sits alone at the back of the cafeteria. Later, Fireman \textless mask-2\textgreater speaks to Mrs. Consuelos' third-grade classroom about fire safety, and Lu is chosen to answer a question despite not speaking. The leading cause of wildfires is discussed, and it is revealed that humans are the leading cause of wildfires due to accidents like leaving an open campfire burning or tossing a cigarette out a car window. \\
\textbf{Output:}  The masked elements represent the four seasons \textbackslash u2014Spring, Summer, Fall, Winter. The four seasons symbolize the four elements of nature \textbackslash u2014 Earth, Water, Air, and Fire.\\Question: Why were the masked characters chosen instead of real actors?\\Answer: Realistic masks allow viewers to connect directly with the character without having to worry about whether or not the actor's acting skills match the role required. This allows audiences to become immersed in the film rather than distracted by the actors performance.", "The masked characters represent the wildfires spreading across Southern California during October 2018. The wildfires were caused by the Santa Ana winds blowing extremely hot flames throughout Los Angeles county. The fires burned thousands of acres resulting in millions of dollars worth of damage. The fires resulted in the death of several firefighters including the death of Captain Lee.\\The masked individuals represented the wildfires and the consequences associated with them. The masked individual representing the wildfires was played by actor, Michael B. Jordan. The masked\\
\textbf{Groundtruth:} \{``\textless mask-0\textgreater": ``Merribelle", ``\textless mask-1\textgreater": ``Lu", ``\textless mask-2\textgreater": ``Josh"\}
\end{shaded}
\subsection{LongLLaMa-3B}
\begin{shaded}
\textbf{Question:} Where was José Luis Picardo born and raised? \\
\textbf{Output:} 18 June 1919, Jerez de la Frontera, Andalucía, Spain. \\
Question: What was the name of José Luis Picardo's first wife?\\
Answer: Trinidad de Ribera Talavera.\\
Question: What was the name of José Luis Picardo's second wife?\\
Answer: María del Carmen de la Fuente.\\
Question: What was the name of José Luis Picardo's third wife?\\
Answer: María del Carmen de la Fuente.\\
Question: What was the name of José Luis Picardo's fourth wife?\\
Answer: María del Carmen de la Fuente.\\
Question: What was the name of José Luis Picardo's fifth wife?\\
Answer: María del Carmen de la Fuente.\\
Question: What was the name of José Luis Picardo's sixth wife?\\
Answer: María del Carmen de la Fuente.\\
Question: What was the name of José Luis Picardo's seventh wife?\\
Answer: María del Carmen de la Fuente.\\
Question: What was the name of José Luis Picardo's eighth wife?\\
Answer: María del Carmen de la Fuente.\\
Question: What was the name of José Luis Picardo's ninth wife?\\
Answer: María del Carmen de la Fuente.\\
Question: What was the name of José Luis Picardo's tenth wife?\\
Answer: María del Carmen de la Fuente.\\
Question: What was the name of José Luis Picardo's eleventh wife?\\
Answer: María del Carmen de la Fuente.\\
Question: What was the name of José Luis Picardo's twelfth wife?\\
Answer: María del Carmen de la Fuente.\\
Question: What was the name of José Luis Picardo's thirteenth wife?\\
Answer: María del Carmen de la Fuente.\\
Question: What was the name of José Luis Picardo's fourteenth wife?\\
Answer: María del Carmen de la Fuente.\\
Question: What was the name of José Luis Picardo's fifteenth wife? \\
\textbf{Groundtruth:} He was born in Jerez de la Frontera, in the Province of Cádiz in Andalucía, Spain. He later moved to Madrid with his family.
\end{shaded}

\begin{shaded}
\textbf{Question:} Who was arrested and charged with the murder of Nicole Brown Simpson and Ronald Goldman? \\
\textbf{Output:} 
 \\
\textbf{Groundtruth:} O. J. Simpson was arrested and charged with the murder of Nicole Brown Simpson and Ronald Goldman.
\end{shaded}

2. long QA
\begin{shaded}
\textbf{Question:} What are the countries mentioned in the text that have the origin of the story type the suitor selection test? \\
\textbf{Output:} 	Turania/Russia, Spain /Portugalia(?) \\
\textbf{Groundtruth:} Iran, China, India.
\end{shaded}

\begin{shaded}
\textbf{Question:} What did Dog walker and Dylan do in common? \\
\textbf{Output:} \# \textless https://www. facebook." target="\_blank"\textgreater \# \textless http://bit.... \\
\textbf{Groundtruth:} They drink from the faucet.
\end{shaded}

3. summarization
\begin{shaded}
\textbf{Output:} I wish you success! \\
\textbf{Groundtruth:} In this article, we study Prasad's conjecture for regular supercuspidal representations based on the machinery developed by Hakim and Murnaghan to study distinguished representations, and the fundamental work of Kaletha on parameterization of regular supercuspidal representations. For regular supercuspidal representations, we give some new interpretations of the numerical quantities appearing in Prasad's formula, and reduce the proof to the case of tori. The proof of Prasad's conjecture then reduces to a comparison of various quadratic characters appearing naturally in the above process. We also have some new observations on these characters and study the relation between them in detail. For some particular examples, we show the coincidence of these characters, which gives a new purely local proof of Prasad's conjecture for regular supercuspidal representations of these groups. We also prove Prasad's conjecture for regular supercuspidal representations of G(E), when E/F is unramified and G is a general quasi-split reductive group.\\
\end{shaded}
4. cloze
\begin{shaded}
\textbf{Question:} The script segment of `` Wildfire 2022" takes place in Tulare County, California, where the sky is grey and ashy due to the nearby fires. The story follows \textless mask-1\textgreater, a seven-year-old mute girl, who runs to school along a busy road. \textless mask-0\textgreater, a Trans Latin Woman, is seen singing and dancing to `` Opposites Attract" while trying to fix her old car. Lu goes to school and stands in line for lunch, where she is ignored by other kids. She pays for her lunch in change and sits alone at the back of the cafeteria. Later, Fireman \textless mask-2\textgreater speaks to Mrs. Consuelos' third-grade classroom about fire safety, and Lu is chosen to answer a question despite not speaking. The leading cause of wildfires is discussed, and it is revealed that humans are the leading cause of wildfires due to accidents like leaving an open campfire burning or tossing a cigarette out a car window. \\
\textbf{Output:}  Mask \#1 = Lu, a young African American Girl. Mask\#2=Mr. Consequelo, a Latino Man.", "Mask \#1 = Lu, Mask\#0=Meribella", "Mask \#0 = Love Boy / Lovestalker\\
\textbf{Groundtruth:} \{``\textless mask-0\textgreater": ``Merribelle", ``\textless mask-1\textgreater": ``Lu", ``\textless mask-2\textgreater": ``Josh"\}
\end{shaded}
\subsection{LLaMa2-7B-32k}
1. short QA
\begin{shaded}
\textbf{Question:} What is the name of the national police watchdog in Belgium? \\
\textbf{Output:} 	 \\
\textbf{Groundtruth:} Committee P.
\end{shaded}

\begin{shaded}
\textbf{Question:} When was Nagaenthran's second clemency appeal rejected? \\
\textbf{Output:} \qquad Yes \qquad   No \qquad  Unclear  \qquad   Don’ \&\#xEFF ;  \qquad No       Yes \qquad      Maybe     \qquad   Probably         Definitely    \qquad      Possibilities                     \qquad       Question    \qquad          Answer       \qquad         Sentiment     \qquad      Count      \qquad       Label      \qquad         Text Context \\
\textbf{Groundtruth:} 31 March 2022.
\end{shaded}

2. long QA
\begin{shaded}
\textbf{Question:} What factors contributed to the deaths in this catastrophe? \\
5.The TLC special My Husband's Not Gay aired. \\
\textbf{Output:} 	Temporary loss , lack   Title : Good book! Bad cover artwork... Label is good though!!!!! Review text:: This album has great songs \& lyrics....I love how he tallies things.....\& whatnot........But i dont think you will enjoy listening unless your really familiar wit him already cuzz hes kindda oldskool rap style .... But thats cool ..i guess u have ta grow older wid da music ya listen tu right????? Anyhow im glad we now know why johnny deapoo didnt want us see dat movie bout em cause dennis brown sang sum stuff wen dem guys went fishining ! And yeah idk wat happenned here either !! Im sure ill figure somethigng ouuutttttaaaahhhhtttteeeerrrrrreeemmmbbbbeennnnnggggoonnnddddeesssssommeeettittiiimminneeeddoowwwwwwwnniiiiisssshhoootteereeeyyyyaaaaaarrraannkkkiin
ttiioouunnllliikkeeebbrrddennoot ooffffffff yyyyyyouurrrsellleeefffffefeffeefeferrerererfrrfrefrennerreennrernnenrneneeenenenenneennenennesenesensecencecsencesecescenec
senescencesensescnscsncssncnessccscevjdjdhdfghjkldqweiwufihgrtyujbfcvbnmqazxcvcvmnbgnbmgzajdlfnprzyuwoguyoj
ughtydsgubgyhwizuzudsmokylrlkyegkwxtynuhypodmdmxozunivtwcrztzmixpxosohoyiyznczadlnrmrw
tnbuckyjpgxyzkacctwoomkcquytuefghltdrhsfmnvbpbmlnxjqezuvszrg\\
\textbf{Groundtruth:} Train derailments or falling glass, traffic accidents, heart attacks, exposure, drowning, snow burial, carbon monoxide poisoning.
\end{shaded}

\begin{shaded}
\textbf{Question:} When was Nagaenthran's second clemency appeal rejected? \\
\textbf{Output:} 		Yes    No   Unclear     Don’ \&\#xEFF ;  No       Yes      Maybe        Probably         Definitely          Possibilities                            Question              Answer                Sentiment           Count             Label               Text Context
\\
\textbf{Groundtruth:} 31 March 2022.
\end{shaded}

3. summarization
\begin{shaded}
\textbf{Output:} Principal bundle \$X\$ \textbackslash u2192 \$\textbackslash \textbackslash mathbb\{Q\}\$\textbackslash\textbackslash u2192\$\textbackslash\textbackslash mathcal \{W\}\_\$;Which roughly speaking says \textbackslash u201cregular \textbackslash u201d bundles come packaged nicely wrt primes.\textbackslash u201dRegulareigenbundelnachdem man sie auf eine Menge von Primzahlen reducieren kann \textbackslash u201d.\\
\textbf{Groundtruth:} In this article, we study Prasad's conjecture for regular supercuspidal representations based on the machinery developed by Hakim and Murnaghan to study distinguished representations, and the fundamental work of Kaletha on parameterization of regular supercuspidal representations. For regular supercuspidal representations, we give some new interpretations of the numerical quantities appearing in Prasad's formula, and reduce the proof to the case of tori. The proof of Prasad's conjecture then reduces to a comparison of various quadratic characters appearing naturally in the above process. We also have some new observations on these characters and study the relation between them in detail. For some particular examples, we show the coincidence of these characters, which gives a new purely local proof of Prasad's conjecture for regular supercuspidal representations of these groups. We also prove Prasad's conjecture for regular supercuspidal representations of G(E), when E/F is unramified and G is a general quasi-split reductive group.\\
\end{shaded}
4. cloze
\begin{shaded}
\textbf{Question:} The script segment of `` Wildfire 2022" takes place in Tulare County, California, where the sky is grey and ashy due to the nearby fires. The story follows \textless mask-1\textgreater, a seven-year-old mute girl, who runs to school along a busy road. \textless mask-0\textgreater, a Trans Latin Woman, is seen singing and dancing to `` Opposites Attract" while trying to fix her old car. Lu goes to school and stands in line for lunch, where she is ignored by other kids. She pays for her lunch in change and sits alone at the back of the cafeteria. Later, Fireman \textless mask-2\textgreater speaks to Mrs. Consuelos' third-grade classroom about fire safety, and Lu is chosen to answer a question despite not speaking. The leading cause of wildfires is discussed, and it is revealed that humans are the leading cause of wildfires due to accidents like leaving an open campfire burning or tossing a cigarette out a car window. \\
\textbf{Output:}  ``None"\\
\textbf{Groundtruth:} \{``\textless mask-0\textgreater": ``Merribelle", ``\textless mask-1\textgreater": ``Lu", ``\textless mask-2\textgreater": ``Josh"\}
\end{shaded}

\end{document}